\documentclass[iicol,pdflatex,sn-mathphys-num]{sn-jnl}

\usepackage{adjustbox}

\usepackage{amssymb}
\usepackage{amsmath}

\usepackage{hyperref}
\usepackage{graphicx}
\usepackage{amsmath}
\usepackage{amssymb}
\usepackage{subcaption}
\usepackage{xcolor}
\usepackage{comment}
\usepackage{enumitem}
\usepackage{float}
\usepackage{booktabs}
\usepackage{tabularx}
\usepackage{multirow}
\usepackage[percent]{overpic}

\theoremstyle{thmstyleone}%
\theoremstyle{thmstyletwo}%

\theoremstyle{thmstylethree}%

\begin{document}

\title[]{Interpretable Plant Leaf Disease Detection Using Attention-Enhanced CNN}


\author[1]{\fnm{Balram} \sur{Singh}}\email{23mcs104@nith.ac.in}

\author*[1]{\fnm{Ram Prakash} \sur{Sharma}}\email{ram.sharma@nith.ac.in}

\author[2]{\fnm{Somnath} \sur{Dey}}\email{somnath@iiti.ac.in}

\affil[1]{\orgdiv{Department of Computer Science and Engineering}, \orgname{ National Institute of Technology Hamirpur}, \orgaddress{\street{Anu}, \city{Hamirpur}, \postcode{177005}, \state{Himachal Pradesh}, \country{India}}}

\affil[2]{\orgdiv{Department of Computer Science and Engineering}, \orgname{ Indian Institute of Technology Indore}, \orgaddress{\street{Simrol}, \city{Indore}, \postcode{453552}, \state{Madhya Pradesh}, \country{India}}}


\abstract{Plant diseases pose a significant threat to global food security, necessitating accurate and interpretable disease detection methods. This study introduces an interpretable attention-guided Convolutional Neural Network (CNN), CBAM-VGG16, for plant leaf disease detection. By integrating Convolution Block Attention Module (CBAM) at each convolutional stage, the model enhances feature extraction and disease localization. Trained on five diverse plant disease datasets, our approach outperforms recent techniques, achieving high accuracy (up to 98.87\%) and demonstrating robust generalization. Here, we show the effectiveness of our method through comprehensive evaluation and interpretability analysis using CBAM attention maps, Gradient-weighted Class Activation Mapping (Grad-CAM), Grad-CAM++, and Layer-wise Relevance Propagation (LRP). This study advances the application of explainable AI in agricultural diagnostics, offering a transparent and reliable system for smart farming. 

}




\maketitle

\section{Introduction}\label{sec1}

The agriculture industry is essential in maintaining food security worldwide, but different crops suffer from a variety of diseases due to varying weather conditions, posing a major threat to crop yield and quality. These diseases are often caused by factors such as extreme temperatures, microbial infections, and changes in humidity or soil conditions. Farmers have always relied on human examination to identify diseases, which is labour-intensive, prone to mistakes, and ineffective on large farms. As a result, among the most crucial areas of research in smart agriculture is the automation of plant disease identification and classification utilizing various Artificial Intelligence (AI) approaches \cite{Singh2019,Dwivedi2021}. In recent years, Convolutional Neural Networks (CNNs) based models in agriculture have performed well on challenges such as classifying plant diseases. However, despite their effectiveness, CNNs often operate as black boxes, offering limited transparency into how predictions are made, hampering trust and broader adoption.  

In order to tackle this, Explainable AI (XAI) methods have been developed enabling visual interpretation of model decisions through attention maps, GradCAM~\cite{Selvaraju2017}, and LRP~\cite{Bach2015}. While GradCAM and GradCAM++ \cite{Chattopadhay2018} generate class-discriminative localization maps using gradients from the final convolutional layers, they may lack resolution and be susceptible to noisy activations. LRP, in contrast, provides pixel-level attributions by backpropagating the model output through a set of layer-specific relevance propagation rules. This offers a more granular explanation of predictions, which is crucial in medical and agricultural diagnostics. 

In this work, an explainable deep learning approach is proposed for the detection of plant leaf disease. Our architecture is based on the VGG16 \cite{Simonyan2014} backbone enhanced with the CBAM~\cite{woo2018cbamconvolutionalblockattention} which introduces attention layers for inherent interpretability of the model with an emphasis on the most relevant features at both the channel and spatial levels. Following each of the five convolutional layers, CBAM modules are added to improve classification accuracy and localization of relevant features by capturing both spatial and channel-wise attention. Five distinct datasets are used to train the model, namely Apple, Plant Village, Embrapa, Maize and Rice to ensure the generalizability and applicability of our proposed method across diverse set of crops. Apart from the inherent interpretability of the proposed method's decision-making process through CBAM layers, we also demonstrate interpretability using advanced explainability methods like LRP, Grad-CAM, and Grad-CAM++. We have also employed high-dimensional feature visualization techniques such as t-distributed Stochastic Neighbor Embedding (t-SNE) and Uniform Manifold Approximation and Projection (UMAP) to visualize feature in lower dimension for visualization of extracted features. Overall, this study advances the application of XAI for agricultural use by presenting an interpretable and performance-robust framework for plant disease classification. 
The following are thes main contributions of the proposed work.
\begin{itemize}
    \item A novel inherently interpretable technique, CBAM-VGG16, is proposed to identify plant leaf disease in various types of crops.
    \item The proposed method enables explainability through feature-level attention across all five convolutional stages, which can be visualized using CBAM attention maps.
    \item Post-hoc explainability of the proposed method is demonstrated using multiple techniques, including Grad-CAM, Grad-CAM++, and a suite of LRP variants.
    \item Feature space analysis using UMAP and t-SNE is conducted to visualize class-wise clustering and inter-model differences.
    \item Superior performance as compared with other state-of-the-art methods on all five datasets demonstrates the generalization and adaptive capability of proposed method for any crop leaf disease identification.
\end{itemize}

\section{Related Works}
\label{related_works}

Barbedo \cite{Barbedo2018} applied CNN architectures to analyze how variations in samples and dataset size influence the efficiency of identifying plant diseases via transfer learning. For this purpose, a dataset comprising 1,383 background-free images across 56 distinct disease types was compiled.
Singh et al. \cite{Singh2019} developed a Multilayer Convolutional Neural Network (MCNN) to classify mango leaves infected by Anthracnose disease. The suggested MCNN model demonstrated higher classification accuracy of 97.13\% compared to other state-of-the-art approaches. 
Dwivedi et al. \cite{Dwivedi2021} have put out the Grape Leaf Disease Detection Network (GLDDN) approach. Their proposed network uses Residual Convolution Neural Network (R-CNN) and multitask learning to identify esca, isariopsis, and black-rot in grape photos. Researchers in \cite{CHEN2021} \cite{Chen2021SEMobileNet} applied different attention strategies within well-known architectures such as MobileNet, Inception, and residual CNNs to identify crop diseases. While these approaches delivered strong results, their evaluation focused on only a limited selection of plant diseases, unlike broader models addressing multiple crops. Cap et al. \cite{Cap2022} has proposed LeafGAN architecture for the augmentation of diseased leaf images via transformation for improving the plant disease diagnosis system using large scale dataset. To solve the data imbalance problem of health vs unhealthy images Zhao et al.\cite{Zhao2022} have used a DoubleGAN architecture. Their architecture uses Wasserstein Generative Adversarial Networks (WGAN) and  Superresolution Generative Adversarial Network (SRGAN) to balance the dataset. Some of the recent methods also used the fuzzy rank-based ensemble along with the pre-trained CNNs \cite{Medjadba2025} and fuzzy feature extraction \cite{Nagi2023} for the plant leaf detection.  

\begin{figure*}[t]
    \centering
    \includegraphics[width=0.9\textwidth]{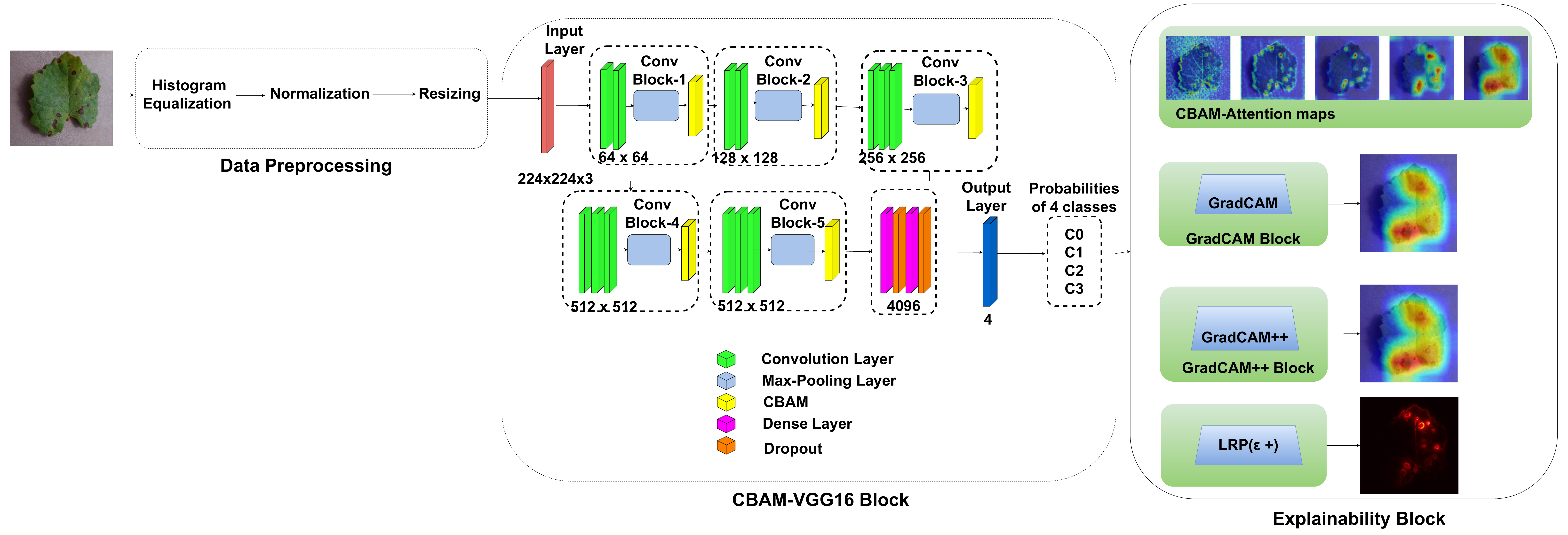}
    \caption{System overview of plant disease classification using CBAM-VGG16.}
    \label{fig:system_overview_1}
\end{figure*}

Although methods like GradCAM \cite{Selvaraju2019} and LRP are commonly utilised to interpret CNN outputs, their application in detecting plant leaf diseases is still relatively new. 
In the context of plant disease identification, LRP has been utilised to highlight significant regions of a leaf, aiding in the detection of disease symptoms and affected zones \cite{Tariq2024}. To improve interpretability, a number of studies have looked into combining GradCAM using additional methods of explanation with the goal of addressing GradCAM’s shortcomings, including its tendency to produce noisy or vague visual outputs. Despite progress, issues like unclear GradCAM explanations and the complexity of LRP visualisations remain unresolved. 
A detailed survey of recent advances in plant disease detection is provided by Qadri et al. \cite{Qadri2025}. They have highlighted the issue and challenges in plant leaf disease detection using machine learning and deep learning based solutions.

\section{Proposed Methodology}
\label{proposed_methodology}

The overview of interpretable architecture used for identifying plant leaf disease is provided in Fig.~\ref{fig:system_overview_1}. The input plant images undergo through data preprocessing to improve quality and guarantee alignment with the input structure of the proposed CBAM-VGG16 model. This involves Contrast Limited
Adaptive Histogram Equalization (CLAHE) to improve contrast, normalization of pixel values to the \([0, 1]\) range and resizing images to \(224 \times 224\) pixels. Following preprocessing, the enhanced VGG16 \cite{Simonyan2014} model integrated with CBAM is employed for disease classification. In this architecture, CBAM modules are added following every convolutional stage of the VGG16 model. The CBAM mechanism adaptively refines the feature maps by applying both channel and spatial attention, thereby enhancing the model’s ability to focus on disease-affected regions. Once the input image's class label is predicted by the model, multiple explainability modules are used to interpret its decision-making process. CBAM attention maps, generated by the integrated CBAM modules in the VGG16 architecture, provide insights into the spatial and channel-wise focus of the network across layers. For class-discriminative localization, Grad-CAM \cite{Selvaraju2019} and Grad-CAM++ \cite{Chattopadhay2018} are used to generate class-discriminative heatmaps that draw attention to the input image's most significant areas. 
Additionally LRP is employed using multiple propagation rules i.e. including $\epsilon$-rule, $\epsilon$-$\gamma$-box, $\alpha_2\beta_1$, and excitation backpropagation to assign pixel-level relevance scores, further enriching the interpretability of the model’s predictions through diverse attribution perspectives.
This entire pipeline from data preprocessing to classification and visual explanation forms an end-to-end plant leaf disease detection system designed for improved interpretability and trust in AI-based agricultural diagnostics. The internal modules of the architecture are explained in further subsections.

\subsection{CBAM VGG16 Architecture}
As illustrated in Fig.~\ref{fig:system_overview_1}, the proposed CBAM-VGG16 architecture begins by utilising input layer of size $224 \times 224 \times 3$. In the architecture, CBAM layers are added after each MaxPooling2D layer. Even though CBAM is a lightweight architecture, overuse could lead to overfitting, which is why the CBAM layer is added after each MaxPooling layer. The channel module and the spatial module are the two primary parts of the CBAM developed  by \cite{woo2018cbamconvolutionalblockattention}. Given the input $I_f \in \mathbb{R}^{c \times W \times H}$, a 1D channel attention map is generated by CBAM  $C_m \in \mathbb{R}^{c \times 1 \times 1}$ as well as a 2D spatial attention map $S_m \in \mathbb{R}^{c \times W \times H}$, where $c$ in our scenario stands for three channel input, $W$ and $H$ stand for the input feature map's width and height, respectively. Our suggested CBAM VGG16's attention process is given in Eqs. 1 and 2.
\begin{align}
    I_c f' &= C_m(I_f) \otimes I_f \tag{1} \\
    I_s f'' &= S_m(I_c f') \otimes I_c f' \tag{2}
\end{align}

where $\otimes$ denotes element-wise multiplication, $I_c f'$ is the channel multiplied output, and $I_s f''$ denotes the final refined output. The internal workings of CBAM's channel attention and spatial attention mechanisms are detailed in the following subsections.



\subsubsection{Channel Attention Module}
A map of channel attention $C_m \in \mathbb{R}^{c \times 1 \times 1}$ is applied to the input features by the channel attention module. To generate the average pooled feature, $I_f^{\text{avg}}$, and the max pooled feature, $I_c^{\text{max}}$, the average and max pooling procedures are applied independently. Thereafter, a shared Multi Layer Perceptron (MLP) network with one hidden layer receives $I_c^{\text{avg}}$ and $I_c^{\text{max}}$. In our CBAM VGG16 design, the activation size of the hidden layer is set to $\mathbb{R}^{c / r \times 1 \times 1}$, where the reduction ratio denoted by $r$, which is set to eight.  This value provides an equitable trade-off between model complexity and representational capacity, as empirically validated in \cite{woo2018cbamconvolutionalblockattention}, ensuring sufficient channel interdependencies are captured without incurring significant computational overhead. The channel attention method used in our proposed architecture is provided as given in Eqs. 3 and 4.

{\footnotesize{
\begin{align}
\footnotesize
C_m(I_f) &= \sigma\big(\text{MLP}(\text{AvgPool}(I_f)) + \text{MLP}(\text{MaxPool}(I_f))\big) \tag{3}
\end{align}}}

{\footnotesize{
\begin{align}
\footnotesize
C_m(I_f) &= \sigma\big(W_1(W_0(I_c^{\text{avg}})) + W_1(W_0(I_c^{\text{max}}))\big) \tag{4}
\end{align}}}



In the above equations, $\sigma$ denotes the sigmoid activation function. The weights $W_0 \in \mathbb{R}^{c \times c/r}$ and $W_1 \in \mathbb{R}^{c/r \times c}$ are shared parameters of the MLP. To generate the channel attention map $C_m(I_f)$, the input feature map $I_f$ undergoes both average pooling and max pooling operations across its spatial dimensions, resulting in $I_c^{\text{avg}}$ and $I_c^{\text{max}}$, respectively. These are then passed through the shared MLP, and their outputs are summed and activated using the sigmoid function. 

\subsubsection{Spatial Attention Module}
This module generates a 2D spatial attention map $S_m \in \mathbb{R}^{c \times W \times H}$, which is applied to the channel-refined feature map $I_c f'$. To compute spatial attention, the module first applies average pooling and max pooling operations along the channel axis, resulting in two 2D feature maps: $I_s^{\text{avg}} \in \mathbb{R}^{1 \times W \times H}$ and $I_s^{\text{max}} \in \mathbb{R}^{1 \times W \times H}$. These two maps are then concatenated along the channel dimension and passed through a convolutional layer to produce the spatial attention map. The resulting map is used to refine the features by performing an element-wise multiplication with $I_c f'$, yielding the final spatially-refined output $I_s f''$. The spatial attention computation in CBAM is formally described in Eqs.~5 and ~6.

{\footnotesize{
\begin{align}
\footnotesize
S_m(I_c f') &= \sigma\big(\nu_{7 \times 7}([\text{AvgPool}(I_c f'); \text{MaxPool}(I_c f')])\big) \tag{5}
\end{align}}}

{\footnotesize{
\begin{align}
\footnotesize
S_m(I_c f') &= \sigma\big(\nu_{7 \times 7}([I_s^{\text{avg}}; I_s^{\text{max}}])\big) \tag{6}
\end{align}}}




In the above equations, $\sigma$ represents the sigmoid activation function, and $\nu_{7 \times 7}$ denotes a convolution operation with a kernel size of $7 \times 7$. 


The intermediate layers of the model utilize the Rectified Linear Unit (ReLU) activation function, defined as given in Eq. 7.

\begin{equation}
    f(x) =
    \begin{cases} 
        x, & \text{if } x > 0 \\
        0, & \text{otherwise} 
    \end{cases} \tag{7}
\end{equation}


Here, $x$ is the input to the activation function. If $x > 0$, the output remains $x$; otherwise, the output is zero. ReLU is preferred over sigmoid and tanh due to its ability to mitigate the vanishing gradient problem. It also aids in learning complex, non-linear patterns and accelerates convergence during training by maintaining sparse activation. 

The final output layer employs the Softmax activation function, a normalized exponential function commonly used for multi-class classification. Softmax transforms the input vector into a probability distribution where the sum of all output probabilities is one, as defined in Eq.~8.


\begin{equation}
\sigma(\vec{z}_i) = \frac{e^{z_i}}{\sum_{j=1}^k e^{z_j}} \tag{8}
\end{equation}
 In our work, as indicated by Eq. 9, we have utilised the Cross Entropy loss function, for model optimization.

\begin{equation}
Loss = -\sum_{i=1}^n y_i \log(\hat{y}_i)\tag{9}
\end{equation}

To reduce the model's complexity, L2 regularization is employed where extreme changes in weights during the training phase are penalized by regularization reducing the probability of over-fitting. During training, the L2 regularizer additionally simplifies the input features and stabilises performance. The definition of the L2 regularizer utilised in our CBAM VGG16 is given in Eq. 10.

\begin{equation}
    R_2 = \frac{\lambda}{2N} \sum_{i=1}^n \|\omega_i\|^2 \tag{10}
\end{equation}

where $\omega_i$ denotes the two-dimensional weight matrix of the $i$th layer, $N$ is the number of input samples, and $\lambda$ is the regularization hyperparameter.

\subsection{Explainability Techniques}
\label{explanation_methods}

\subsubsection{Post-hoc Attribution Methods}
Methods such as Grad-CAM \cite{Selvaraju2019} and Grad-CAM++ \cite{Chattopadhay2018}, utilize the gradients of class scores with respect to intermediate feature maps to produce localization maps. Grad-CAM is a widely-used technique to show the areas of an input image that are distinctive to a class and have the biggest influence on a model's judgment. Grad-CAM++ is a generalized version of Grad-CAM that enables more accurate localization, especially in scenarios with several instances of the same item.

\subsubsection{LRP Techniques}
LRP~\cite{Bach2015}, \cite{Montavon2017} belongs to the class of additive explanation techniques. These methods operate under the assumption that a function $f_j$ with $N$ input features $x = \{x_i\}_{i=1}^N$ can be expressed as a sum of contributions from each input variable, represented as $R_{i \leftarrow j}$, referred to as \emph{relevance scores}. Here, $R_{i \leftarrow j}$ quantifies how much the $i$th input contributes to the $j$-th output. The total function value can be approximated (or exactly recovered) as shown in Eq.~\ref{eq:lrp1}.

\begin{equation}
f_j(x) \propto R_j = \sum_i R_{i \leftarrow j}
\tag{11} \label{eq:lrp1}
\end{equation}

When an input $x_i$ influences multiple outputs $j$, as is common in multidimensional functions, the total relevance attributed to $x_i$ is the sum over its contributions from each output, as defined in Eq.~\ref{eq:lrp2}.

\begin{equation}
R_i = \sum_j R_{i \leftarrow j}
\tag{12} \label{eq:lrp2}
\end{equation}

Unlike other interpretability approaches, LRP explicitly treats the neural network as a hierarchical, layer-wise acyclic graph where each unit $j$ in layer $l$ is associated with a local function $f^l_j$. Relevance values from the output layer $L$, where $R^L_j \propto f^L_j$, are passed backward, layer by layer, through the network until it reaches the input. This reverse mapping follows the same activation path used during the forward inference process, moving from the output node $f^L$ down to the initial input $f^1$.




We have also used some of the advanced variants of LRP such as Epsilon Plus Rule (\(\varepsilon^+\))\cite{Bach2015}, Epsilon Plus Gamma Box Rule (\(\varepsilon^+ \gamma \boxdot\)) \cite{anders2023softwaredatasetwidexailocal}, Epsilon Plus Flat Rule (\(\varepsilon^+_{\text{flat}}\))\cite{Montavon2017}, and Epsilon Alpha2 Beta 1 Flat Rule (\(\varepsilon^{\alpha=2,\beta=1}_{\text{flat}}\))\cite{anders2023softwaredatasetwidexailocal} for assesing the interpretability of our proposed method.

\section{Experimental Results}
\label{results_and discussion}

\subsection{Datasets}  
We have selected five different datasets for to evaluate our proposed method. These five datasets, namely, PlantVillage, Embrapa, Maize, Apple, and Rice ensure the generalizability of our approach in plant leaf disease detection for other crops. All the datasets are divided in an almost 80:20 ratio in the training and testing set. The overall composition of the datasets is provided in Table \ref{tab:datasets}.

\begin{table}[t]
\caption{Description of different datasets used in our work.}
\label{tab:datasets}
\centering
\tiny
\setlength{\tabcolsep}{1pt}
\begin{tabularx}{1\columnwidth}{lccccc}
\hline
\textbf{Dataset} & \textbf{\begin{tabular}[c]{@{}c@{}}Total\\ Images\end{tabular}} & \textbf{Training} & \textbf{Testing} & \textbf{\begin{tabular}[c]{@{}c@{}}Total\\ Disease\end{tabular}}& \textbf{\begin{tabular}[c]{@{}c@{}} Disease/Crops\end{tabular}} \\
\hline
PlantVillage & 54,303 & 43,442 & 10,861 & 38 & \begin{tabular}[c]{@{}c@{}}Apple Scab, Grape \\ Esca, Tomato Early\\ Blight, etc.\end{tabular}  \\
\hline
Embrapa & 46,376 & 37,100 & 9,276 & 93 &  \begin{tabular}[c]{@{}c@{}}Coffee, Cotton, \\ Soybean, Citrus, \\Grapevine etc. \end{tabular}\\
\hline
Maize & 3,852 & 3,082 & 770 & 4 & \begin{tabular}[c]{@{}c@{}}Gray Leaf Spot, \\ Northern Leaf \\ Blight etc.\end{tabular} \\
\hline
Apple & 3,644 & 2921  & 723 & 4 & \begin{tabular}[c]{@{}c@{}}Cedar Apple Rust, \\ Healthy, Apple Scab etc.\end{tabular} \\
\hline
Rice & 9,000 & 7,200 & 1,800 & 10 & \begin{tabular}[c]{@{}c@{}}Brown Spot, Blast,\\ Tungro etc.\end{tabular}  \\
\hline
\end{tabularx}
\end{table}

\subsection{Evaluation Metrics}  
The F1 score (F1), Area Under Curve (AUC), Recall (REC), Accuracy (ACC), Precision (PREC), and Cohen's Kappa score (KAPPA) are common classification metrics that are used to assess and compare the performance of the proposed CBAM-VGG16 model with state-of-the-art models. The model’s feature representation capability is further assessed through high-dimensional visualization techniques, t-SNE \cite{JMLR:v9:vandermaaten08a} and UMAP \cite{mcinnes2020umapuniformmanifoldapproximation}.


\begin{table}[!h]
\centering
\caption{Comparison of the work with other state-of-the-art methods on five publicly available datasets.}
\label{tab:my-table} 
\tiny
\setlength{\tabcolsep}{2pt}
\begin{tabular}{@{}llllllll@{}}
\toprule
\multicolumn{8}{c}{\textbf{Apple Dataset}}                                                                                                  \\ \midrule
Author                 & LOSS          & ACC            & PREC           & REC            & F1             & AUC            & KAPPA         \\
Karthik et al. \cite{Karthik2020}               & 1.56          & 62.37          & 62.84          & 61.83          & 62.35          & 83.53          & 0.45          \\
Chen et al. \cite{CHEN2021}                   & 0.79          & 83.33          & 83.70          & 82.80          & 83.25          & 94.10          & 0.76          \\
Chen et al. \cite{Chen2021rice}                  & 0.63          & 83.33          & 85.47          & 82.26          & 83.83          & 95.59          & 0.76          \\
Li et al. \cite{Hangli}                     & 1.34          & 41.94          & 44.59          & 35.48          & 39.52          & 71.68          & 0.16          \\
Chen et al. \cite{chenjunde}                   & 0.73          & 87.10          & 87.10          & 87.10          & 87.10          & 95.02          & 0.81          \\
Yang et al. \cite{YANG2023107543}                   & 0.49          & 85.48          & 85.71          & 83.87          & 84.78          & 96.44          & 0.79          \\
Thakur et al. \cite{THAKUR2023}                 & \textbf{0.30} & 93.55          & 93.55          & 93.55          & 93.55          & 97.01          & 0.91          \\
Thakur et al. \cite{PSTHAKUR}             & 0.43          & 94.62          & 94.62          & 94.62          & 94.62          & 98.08          & \textbf{0.92} \\
Tong Li et al. \cite{li2024yololeaf} & -- & 93.89 & 93.84 & -- & 93.80 & -- & -- \\

\textbf{Proposed work} & 0.41 & \textbf{95.42} & \textbf{95.27} & \textbf{95.42} & \textbf{95.27} & \textbf{99.05} & 0.90 \\ \midrule

\multicolumn{8}{c}{\textbf{Embrapa Dataset}}                                                                                                \\ \midrule
Karthik et al. \cite{Karthik2020}                & 0.77          & 80.29          & 83.07          & 78.38          & 80.60          & 97.80          & 0.82          \\
Chen et al. \cite{CHEN2021}                   & 1.11          & 73.63          & 80.12          & 67.86          & 73.48          & 98.02          & 0.73          \\
Chen et al. \cite{Chen2021rice}                   & 1.12          & 74.88          & 82.93          & 66.06          & 73.18          & 98.20          & 0.75          \\
Li et al. \cite{Hangli}                     & 0.90          & 73.17          & 79.23          & 67.59          & 72.95          & 98.38          & 0.72          \\
Chen et al. \cite{chenjunde}                   & 0.28          & 93.19          & 94.33          & 93.47          & 93.90          & 99.08          & 0.94          \\
Yang et al. \cite{YANG2023107543}                   & 0.62          & 88.48          & 90.91          & 86.53          & 88.67          & 99.51          & 0.88          \\
Thakur et al. \cite{THAKUR2023}                 & 0.46          & 89.24          & 91.17          & 88.27          & 89.70          & 98.73          & 0.89          \\
Thakur et al. \cite{PSTHAKUR}             & 0.34          & 93.83          & 94.78          & 93.21          & 93.99          & 99.68          & 0.94          \\
\textbf{Proposed work} & \textbf{0.32} & \textbf{94.20} & \textbf{94.70} & \textbf{93.90} & \textbf{94.30} & \textbf{99.75} & \textbf{0.95} \\ \midrule

\multicolumn{8}{c}{\textbf{Maize Dataset}}                                                                                                  \\ \midrule
Karthik et al. \cite{Karthik2020}               & 1.17          & 54.32          & 59.72          & 53.09          & 56.21          & 83.19          & 0.39          \\
Chen et al. \cite{CHEN2021}                   & 0.60          & 87.65          & 87.65          & 87.65          & 87.65          & 96.03          & 0.84          \\
Chen et al. \cite{Chen2021rice}                   & 0.50          & 88.89          & 91.03          & 87.65          & 87.65          & 96.78          & 0.85          \\
Li et al. \cite{Hangli}                     & 1.38          & 50.62          & 51.47          & 43.21          & 46.98          & 72.47          & 0.34          \\
Chen et al. \cite{chenjunde}                   & 0.54          & 91.36          & 91.36          & 91.36          & 91.36          & 97.29          & 0.89          \\
Yang et al. \cite{YANG2023107543}                   & 1.52          & 76.54          & 78.48          & 76.54          & 77.50          & 94.53          & 0.69          \\
Thakur et al. \cite{THAKUR2023}                 & 0.34          & 92.59          & \textbf{93.67} & 91.36          & 92.50          & 97.21          & 0.90          \\
Thakur et al. \cite{PSTHAKUR}             & \textbf{0.04} & 92.59          & 92.59          & 92.59          & 92.59          & 96.27          & 0.90          \\
Alpsalaz et al. \cite{Alpsalaz2025} & -- & 94.97 & -- & -- & -- & -- & -- \\

\textbf{Proposed work} & 0.12 & \textbf{95.00} & 95.05 & \textbf{95.00} & \textbf{95.01} & \textbf{99.62} & \textbf{0.93} \\ \midrule

\multicolumn{8}{c}{\textbf{PlantVillage Dataset}}                                                                                           \\ \midrule
Karthik et al. \cite{Karthik2020}                & 0.16          & 95.83          & 96.20          & 95.60          & 95.89          & 99.70          & 0.96          \\
Chen et al. \cite{CHEN2021}                   & 1.07          & 74.63          & 80.92          & 69.64          & 74.03          & 98.18          & 0.74          \\
Chen et al. \cite{Chen2021rice}                   & 0.17          & 96.68          & 97.49          & 95.83          & 96.64          & 99.26          & 0.97          \\
Li et al. \cite{Hangli}                     & 0.46          & 86.51          & 88.85          & 85.08          & 86.92          & 99.40          & 0.85          \\
Chen et al. \cite{chenjunde}                   & 0.12          & 97.27          & 97.29          & 97.27          & 97.27          & 99.63          & 0.97          \\
Yang et al. \cite{YANG2023107543}                   & 0.09          & 97.69          & 97.83          & 97.69          & 97.71          & 99.82          & 0.98          \\
Thakur et al. \cite{THAKUR2023}                 & 0.04          & 98.46          & 98.52          & 98.46          & 98.46          & 99.91          & 0.98          \\
Thakur et al. \cite{PSTHAKUR}             & 0.06          & 98.29          & 98.30          & 98.29          & 98.29          & \textbf{99.95} & 0.98          \\

Shafik et al. \cite{shafik2024transfer} & -- & 97.8 & 97.7 & 96.8 & 94.4 & -- & -- \\

\textbf{Proposed work} & \textbf{0.03} & \textbf{98.72} & \textbf{98.72} & \textbf{98.72} & \textbf{98.72} & \textbf{99.95} & \textbf{0.99} \\ \midrule

\multicolumn{8}{c}{\textbf{Rice Dataset}}                                                                                                   \\ \midrule
Karthik et al. \cite{Karthik2020}               & 0.13          & 96.30          & 96.41          & 96.30          & 96.30          & 99.76          & 0.96          \\
Chen et al. \cite{CHEN2021}                   & 0.45          & 85.92          & 89.25          & 85.92          & 85.96          & 98.90          & 0.86          \\
Chen et al. \cite{Chen2021rice}                   & 0.14          & 95.06          & 95.19          & 95.06          & 95.06          & 99.41          & 0.95          \\
Li et al. \cite{Hangli}                     & 0.54          & 91.01          & 92.69          & 91.01          & 91.01          & 99.61          & 0.91          \\
Chen et al. \cite{chenjunde}                   & 0.08          & 97.20          & 97.31          & 97.20          & 97.20          & 99.81          & 0.97          \\
Yang et al. \cite{YANG2023107543}                   & 0.14          & 95.67          & 95.79          & 95.67          & 95.67          & 99.84          & 0.95          \\
Thakur et al. \cite{THAKUR2023}                 & 0.04          & 97.59          & 97.59          & 97.59          & 97.59          & 99.88          & 0.97          \\
Thakur et al. \cite{PSTHAKUR}             & 0.03          & 98.10          & 98.11          & 98.10          & 98.10          & 99.93          & 0.98          \\
Bhuyan et al. \cite{Bhuyan2024} & -- & 98.41 & 98.58 & 98.39 & 98.19 & -- & -- \\

\textbf{Proposed work} & \textbf{0.01} & \textbf{98.87} & \textbf{98.87} & \textbf{98.87} & \textbf{98.87} & \textbf{99.94} & \textbf{0.99} \\ \bottomrule
\end{tabular}
\end{table}

\subsection{Results and Comparative Study}
\label{comparative_analysis}


The performance results along with a detailed comparative study is reported in Table \ref{tab:my-table}. It can be observed from the results that across all five datasets, our CBAM-VGG16 consistently outperforms existing models \cite{Karthik2020},  \cite{CHEN2021}, \cite{Chen2021rice}, \cite{THAKUR2023}, \cite{Hangli}, \cite{chenjunde}, \cite{YANG2023107543}, \cite{PSTHAKUR}, \cite{li2024yololeaf}, \cite{shafik2024transfer}, \cite{Bhuyan2024}, \cite{Alpsalaz2025}. On the Apple dataset, the our method performs better with the highest accuracy of 95.42\%. On the Embrapa dataset, CBAM-VGG16 again achieves best performance, with an accuracy of 94.20\%. The similar superiority  performance is observed in other evaluation metrics as well, indicating the robust generalization capability of our proposed method across different datasets. For the Maize dataset, our model obtains a notable accuracy of 95.00\%, outperforming all the methods. In the PlantVillage dataset, the proposed model maintains a competitive edge with nearly perfect scores across all metrics. Notably, it again exceeds in performance, reaffirming its superior feature learning and attention capabilities as compared with other methods. Lastly, on the Rice dataset, our approach attains the maximum accuracy of 98.87\%, alongside the best precision, recall, and AUC (99.94\%), demonstrating its effectiveness in capturing fine-grained disease patterns even in challenging samples. The obtained results on all the datasets exhibits that the suggested approach has good generalizability and can be adapted to identify leaf diseases in any other crop.

\begin{figure}[t]
    \centering
    \includegraphics[width=1\linewidth]{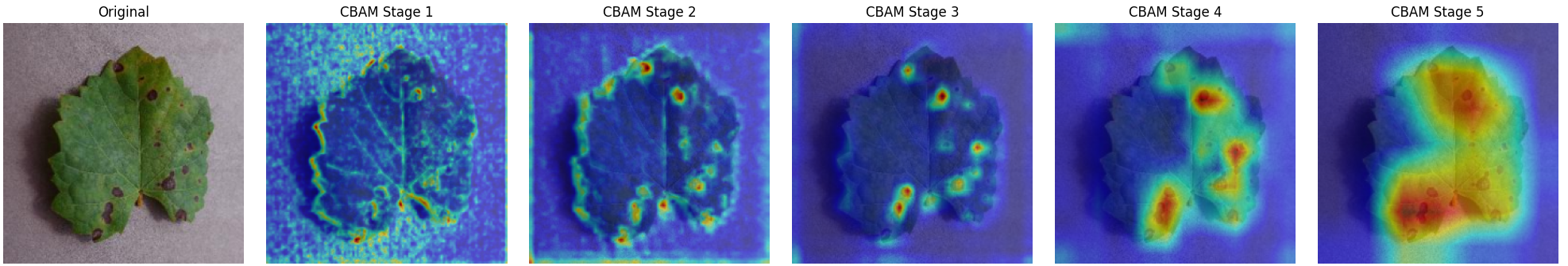}
    \vspace{1mm}
    \includegraphics[width=1\linewidth]{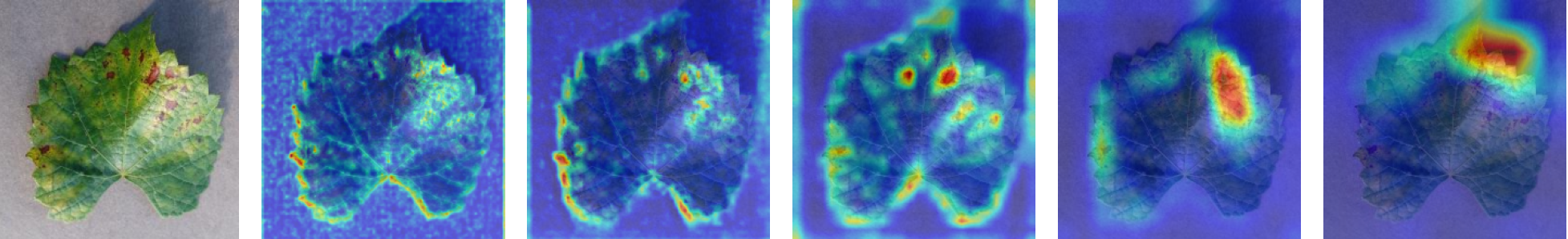}
    
    \includegraphics[width=1\linewidth]{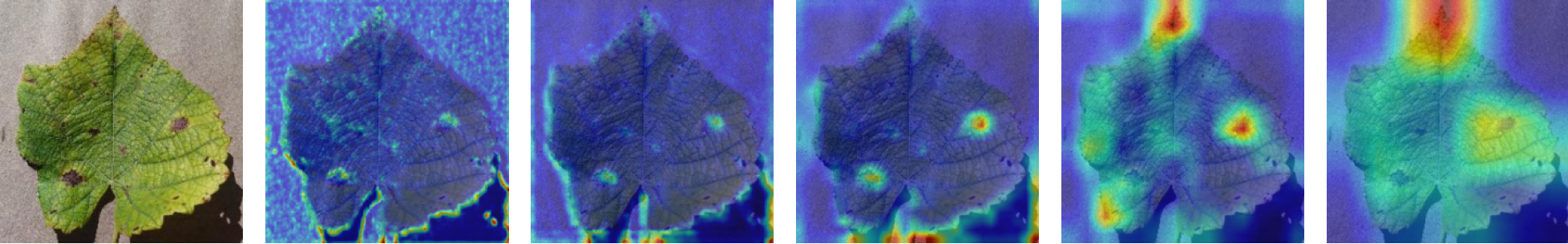}
    \vspace{0.3mm}
    \includegraphics[width=1\linewidth]{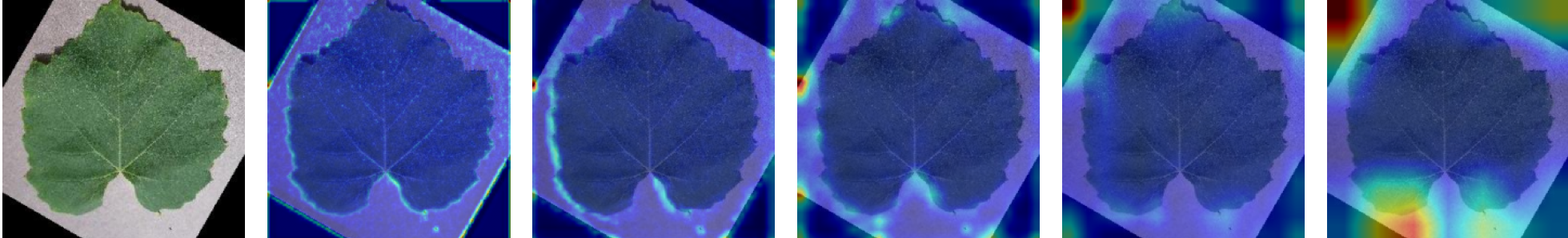}
    \caption{Illustrations of CBAM attention maps generated for inputs of different classes on a CBAM-VGG16. Rows from top to bottom correspond to images of class black rot, esca, leaf blight, and healthy, respectively.}
    \label{fig:grape_leaf_explanations_2}
\end{figure}

  \begin{figure}[h]
    \centering
    \includegraphics[width=0.5\linewidth]{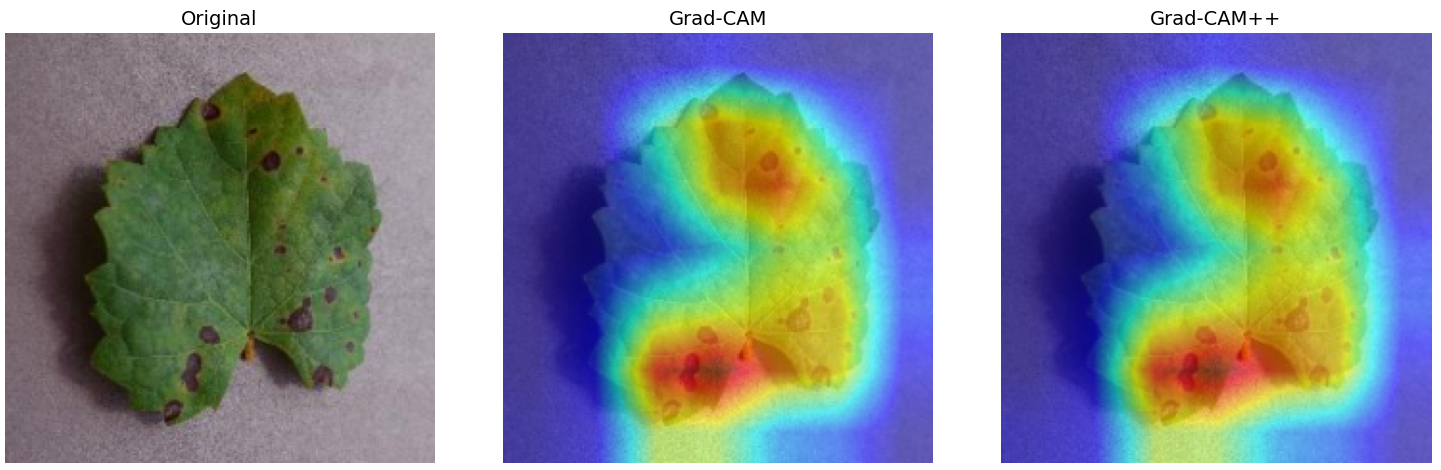}
 
    \includegraphics[width=0.5\linewidth]{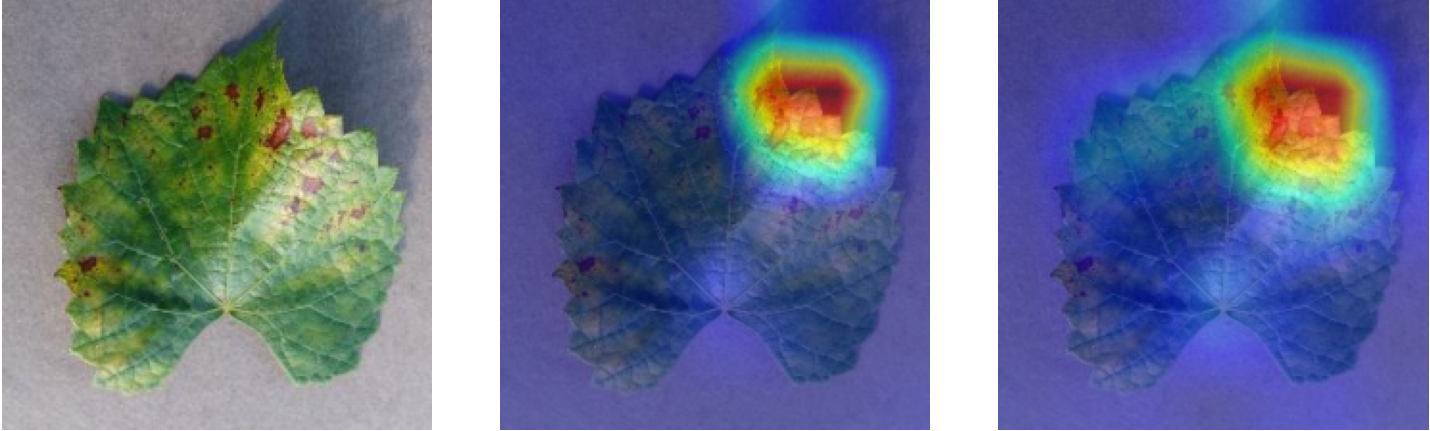}
    
    \includegraphics[width=0.5\linewidth]{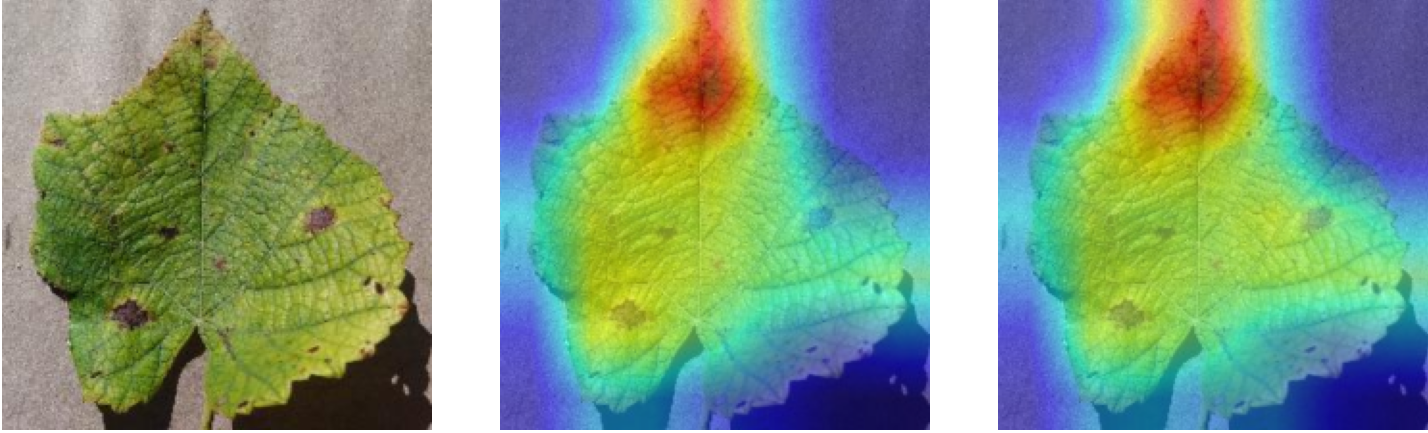}
    
    \includegraphics[width=0.5\linewidth]{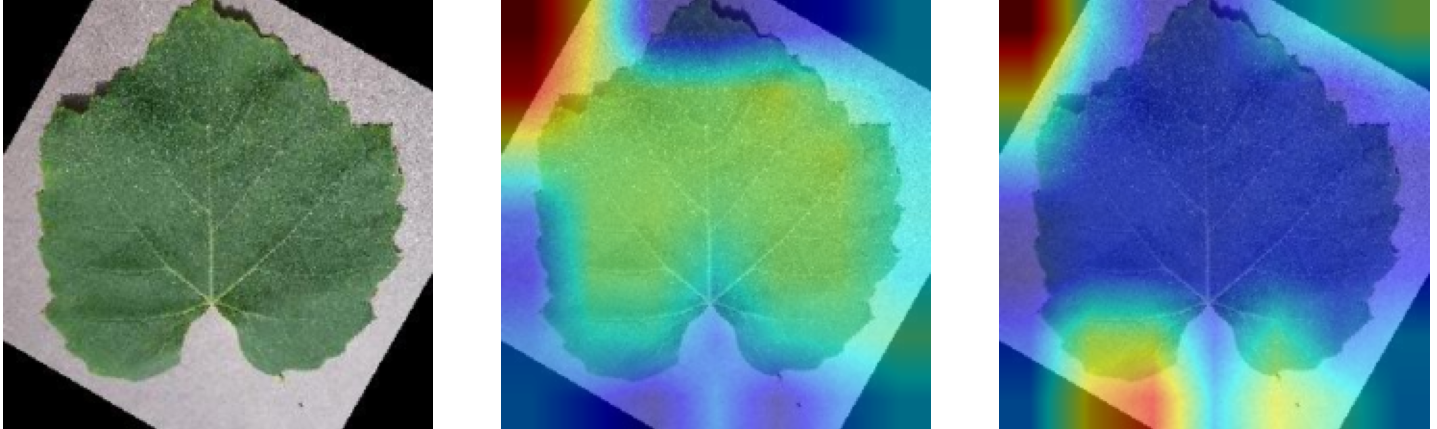}
    
    \caption{Visualization of GradCAM and GradCAM++ outputs. Rows from top to bottom correspond to images of class black rot, esca, leaf blight, and healthy, respectively.}
    \label{fig:grape_leaf_explanations_1}
\end{figure}

\subsection{Explainability Analysis}
\label{explainability_analysis}

\begin{figure}[t]
\centering
\setlength{\tabcolsep}{1pt} 
\renewcommand{\arraystretch}{0.5} 

\begin{tabular}{ccccc}

\includegraphics[width=0.55in]{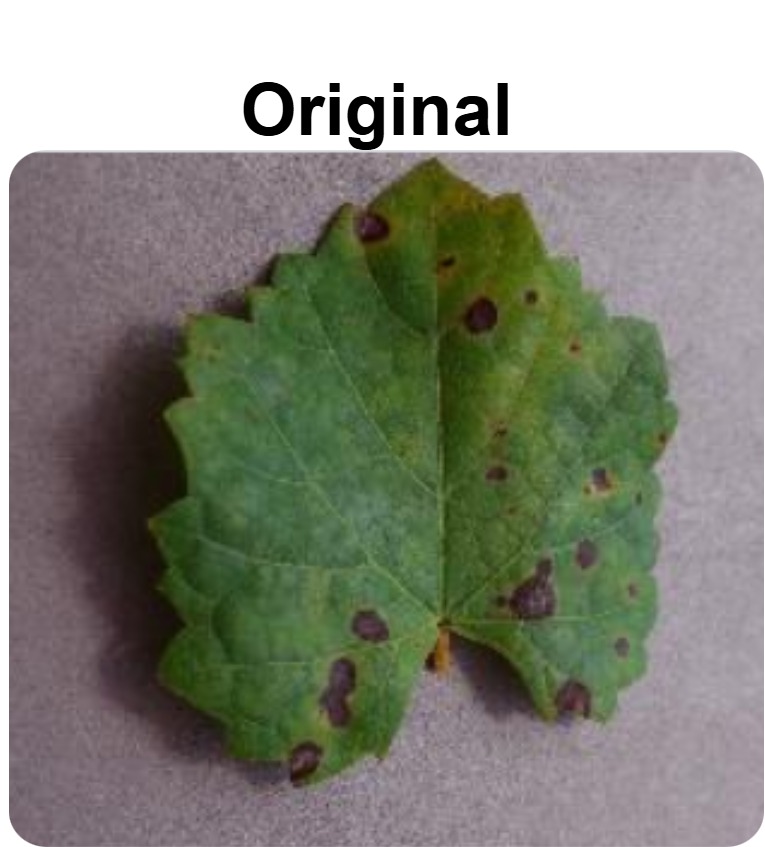} &
\includegraphics[width=0.5in]{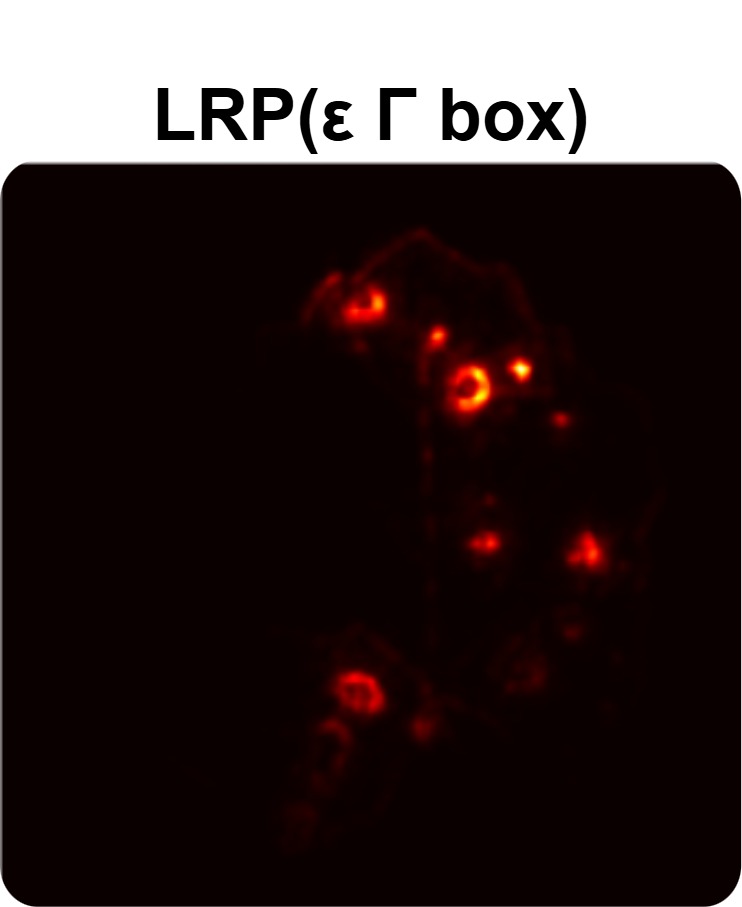} &
\includegraphics[width=0.52in]{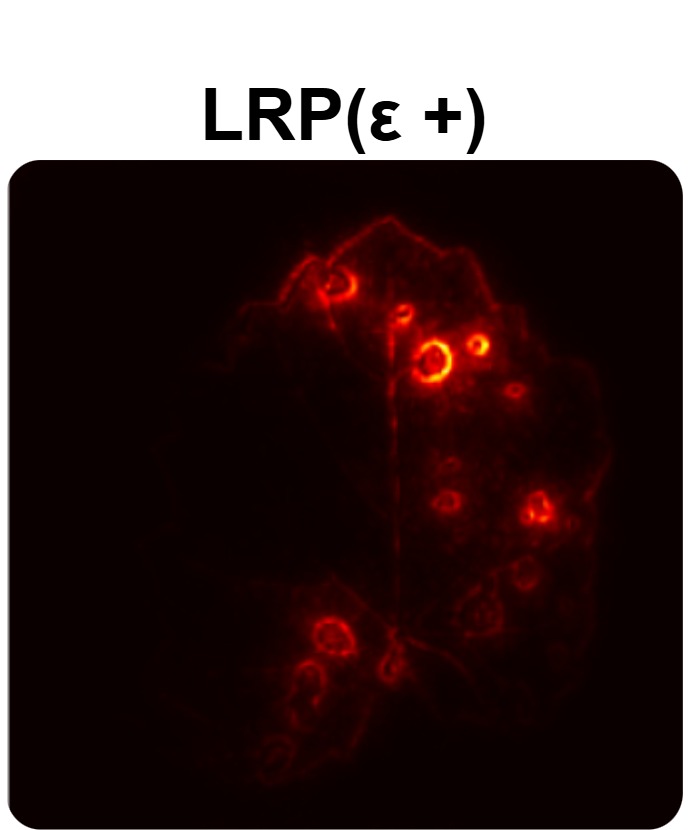} &
\includegraphics[width=0.5in]{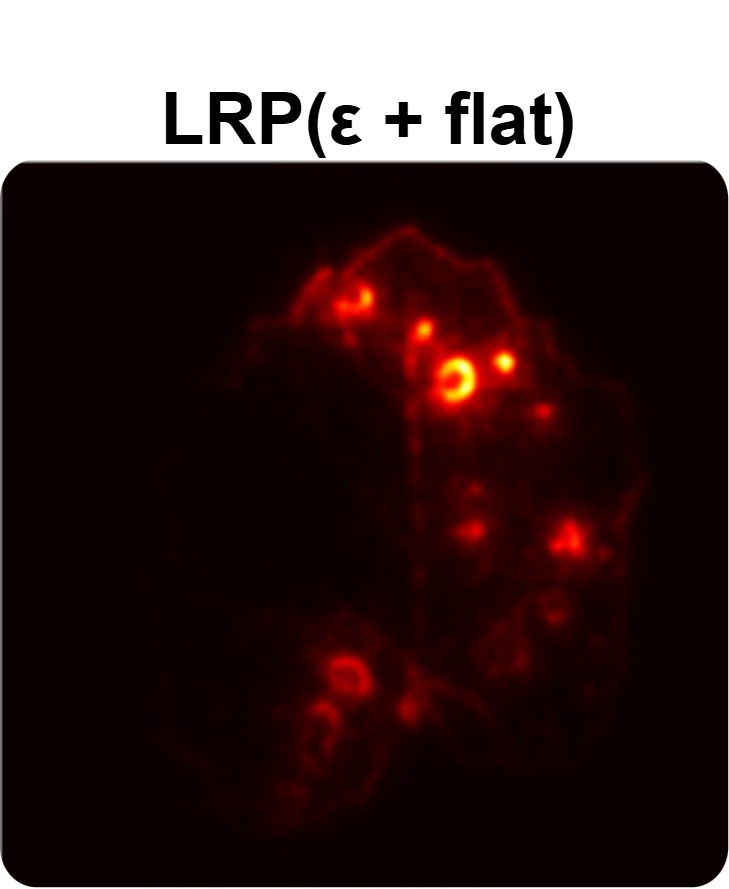} &
\includegraphics[width=0.52in]{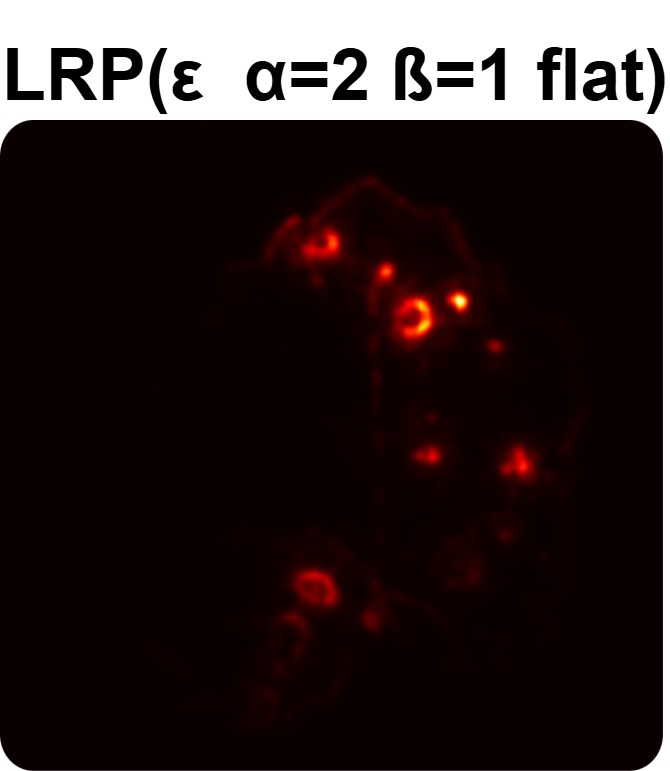} \\

\includegraphics[width=0.55in]{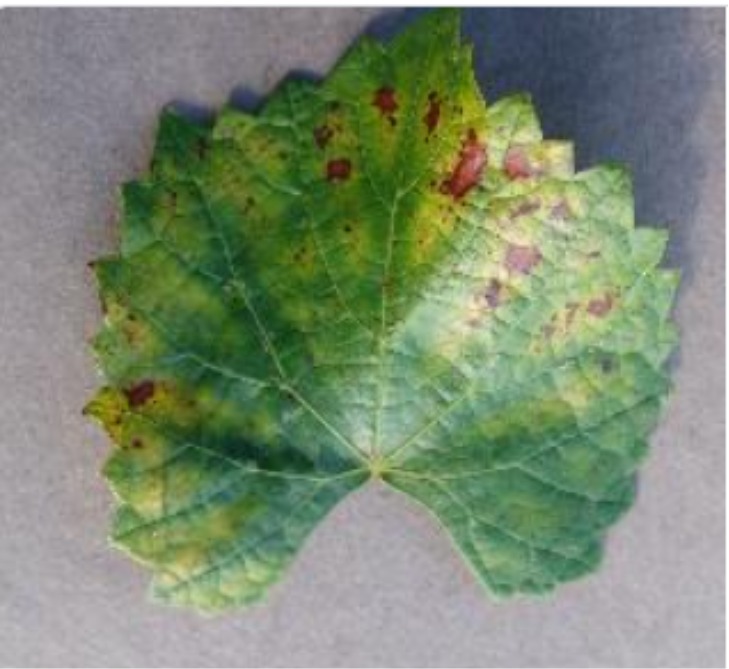} &
\includegraphics[width=0.52in]{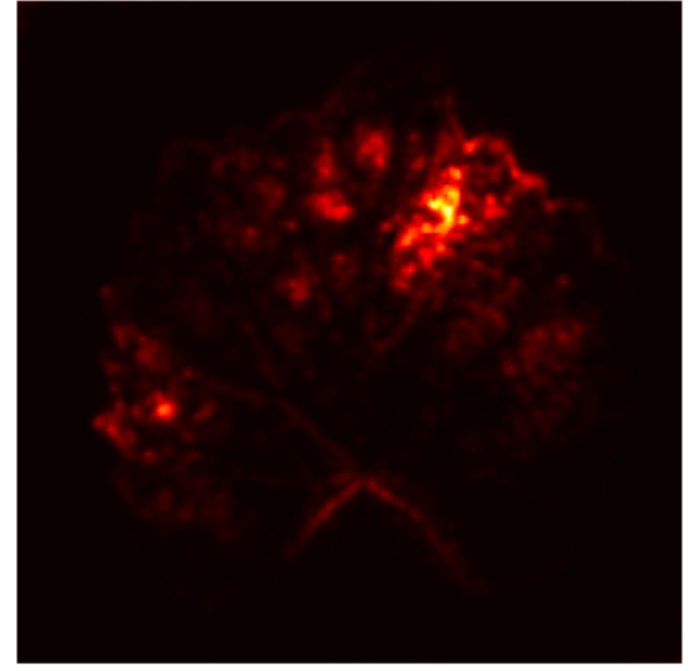} &
\includegraphics[width=0.5in]{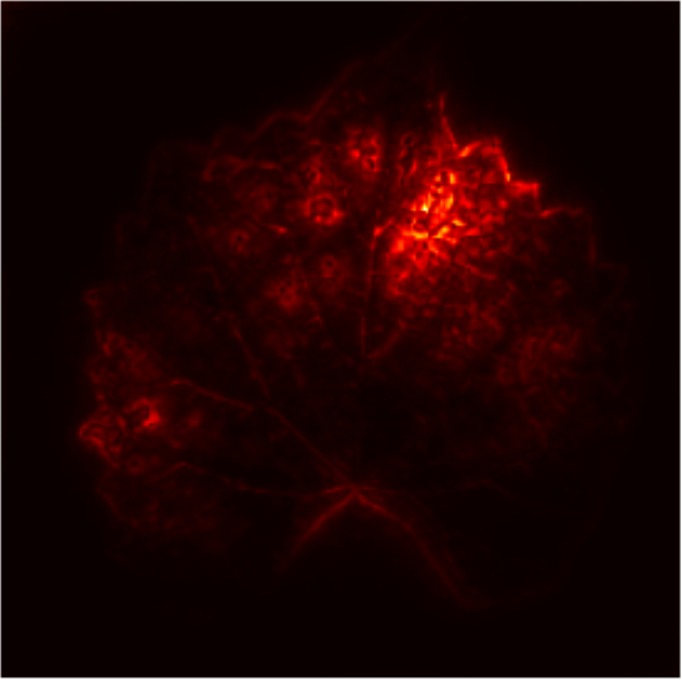} &
\includegraphics[width=0.52in]{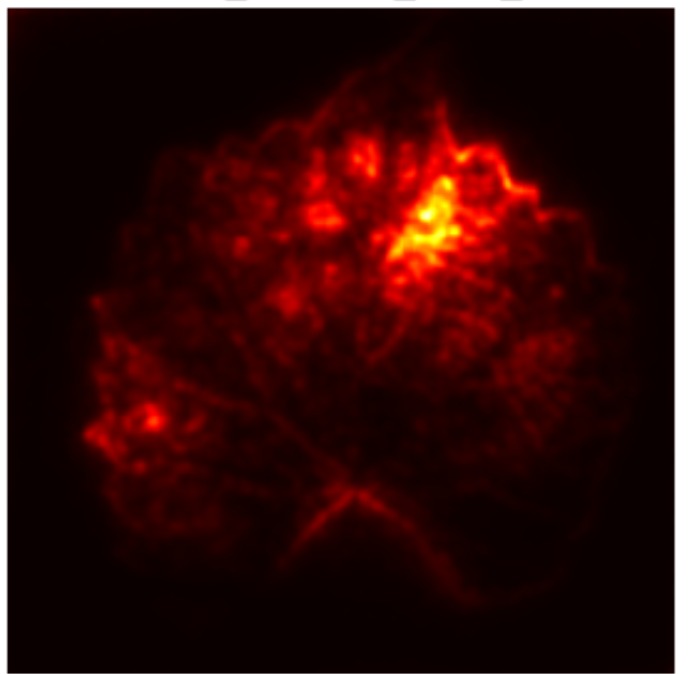} &
\includegraphics[width=0.52in]{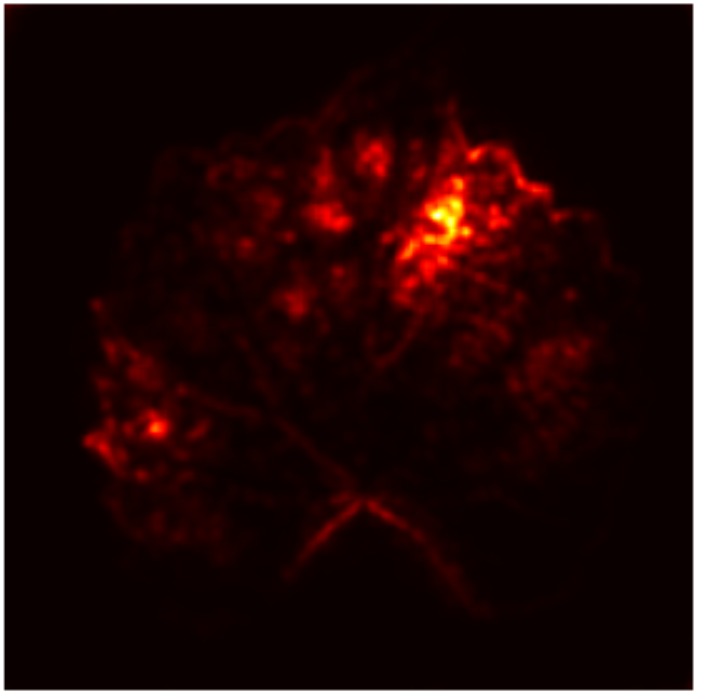} \\

\includegraphics[width=0.55in]{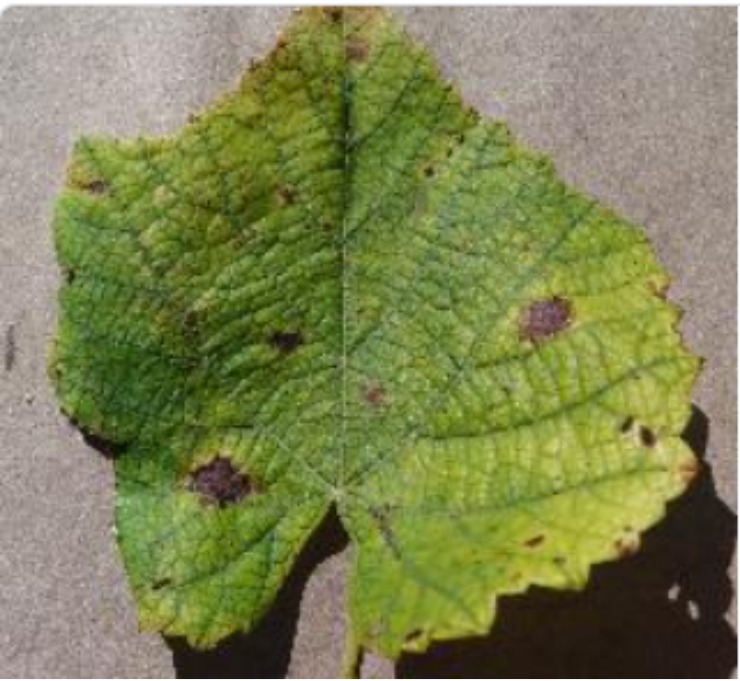} &
\includegraphics[width=0.5in]{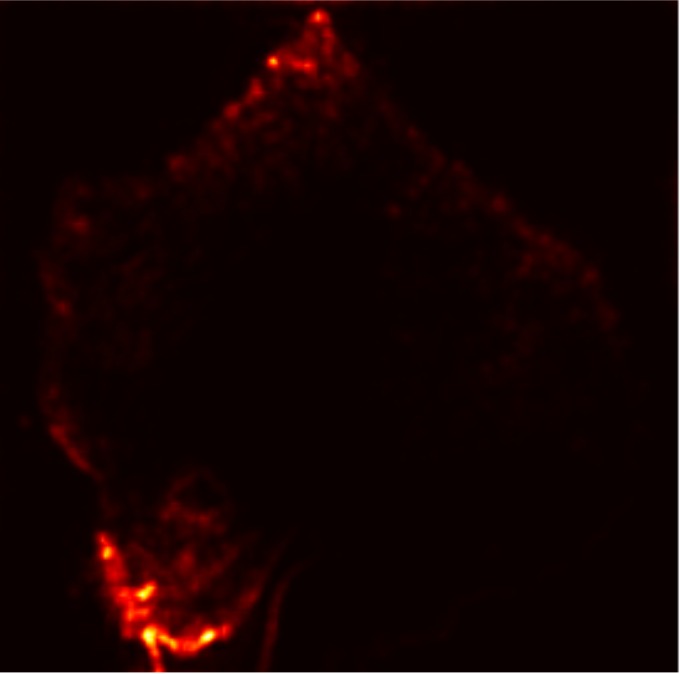} &
\includegraphics[width=0.5in]{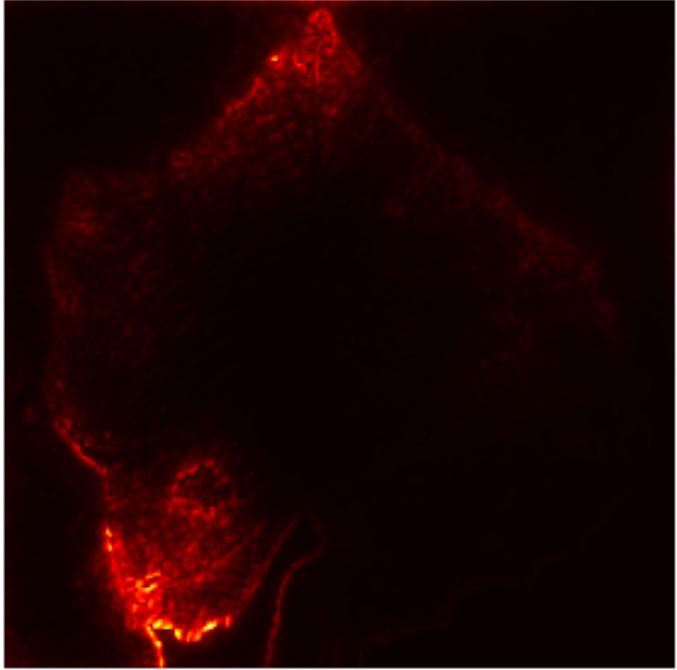} &
\includegraphics[width=0.52in]{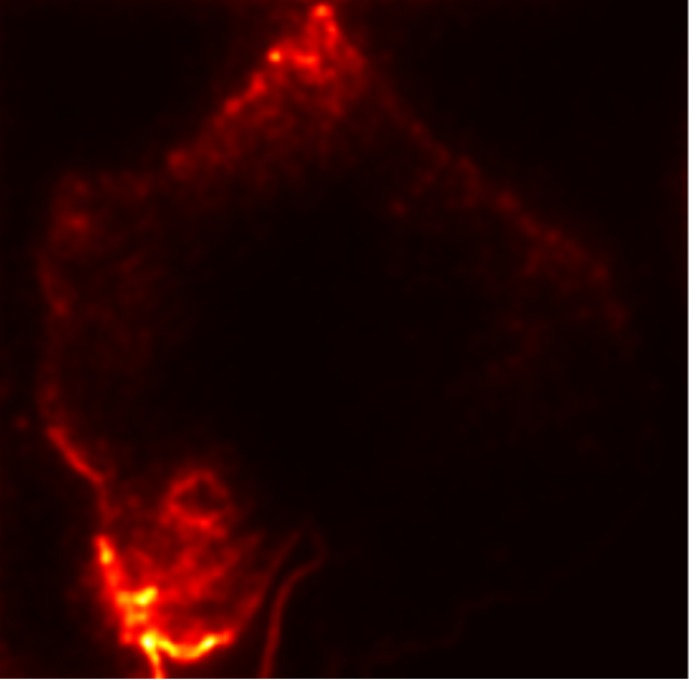} &
\includegraphics[width=0.52in]{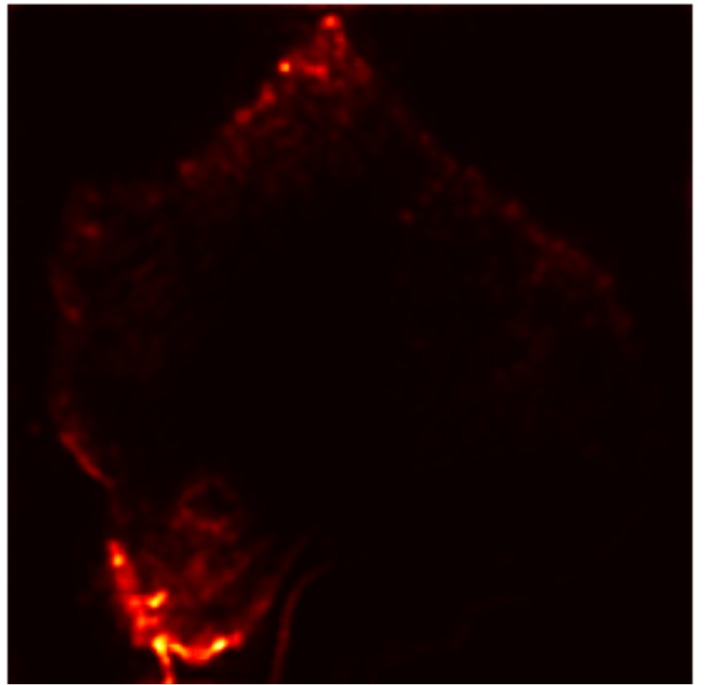} \\

\includegraphics[width=0.55in]{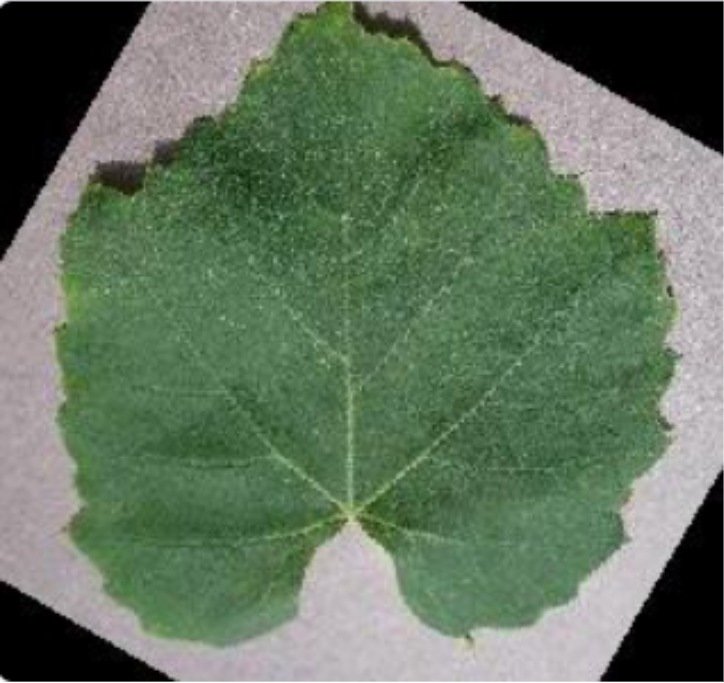} &
\includegraphics[width=0.51in]{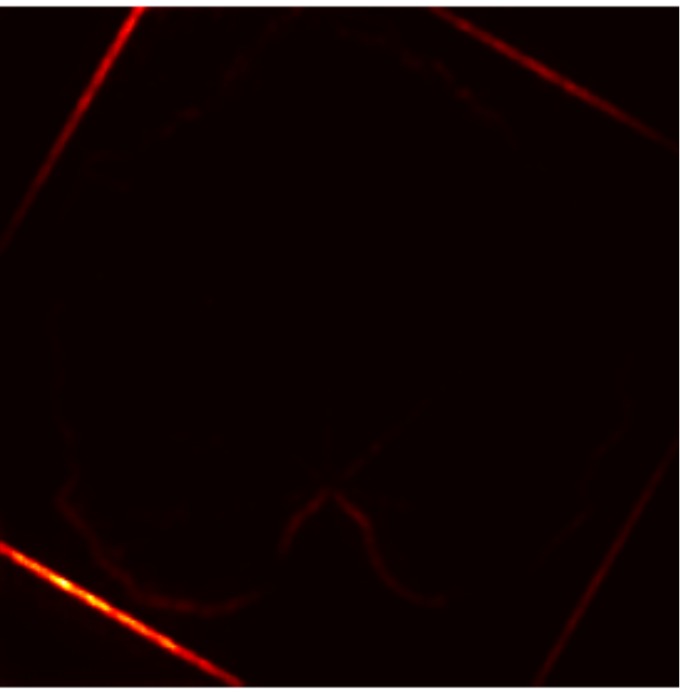} &
\includegraphics[width=0.51in]{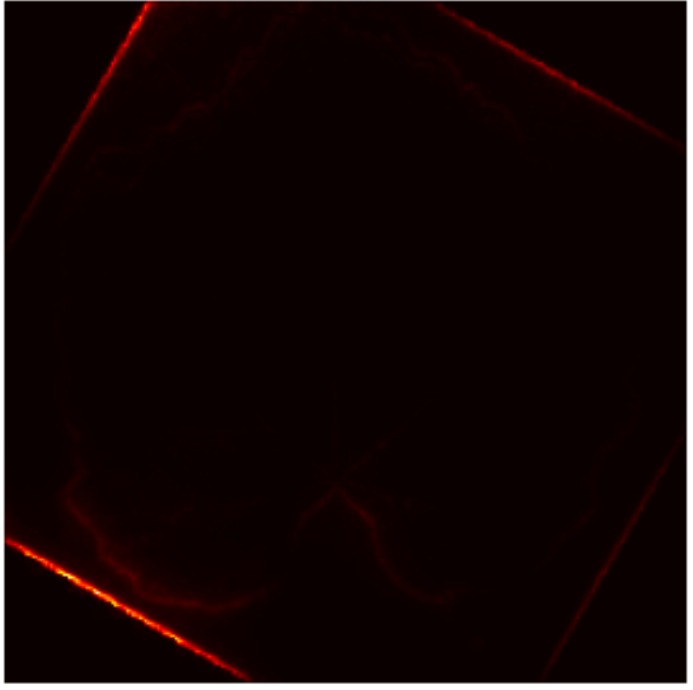} &
\includegraphics[width=0.51in]{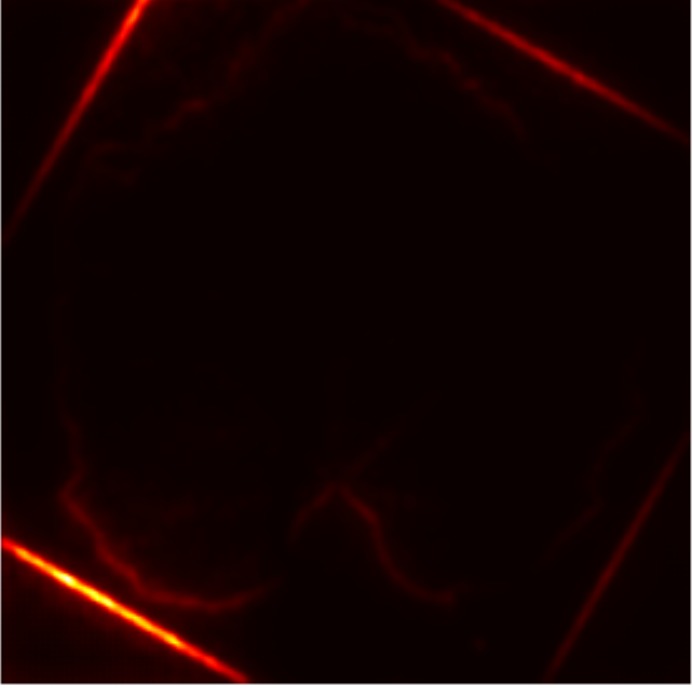} &
\includegraphics[width=0.51in]{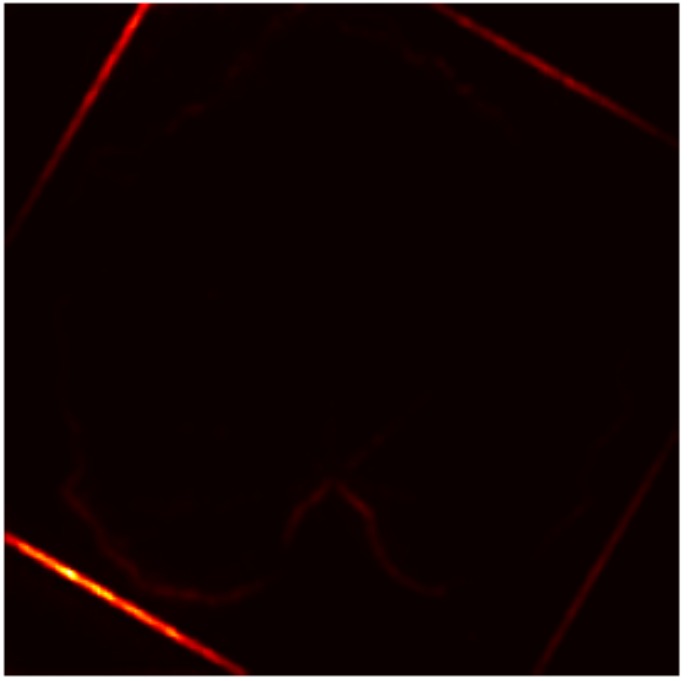} \\

\end{tabular}

\caption{LRP visualization across multiple rules and leaf samples. Each row shows a different test image of class black rot, esca, leaf blight and  healthy, respectively; each column corresponds to a different LRP technique.}
\label{fig:lrp_grid_1}
\end{figure}

\begin{figure*}[!thb]
\centering
\setlength{\tabcolsep}{1pt} 
\renewcommand{\arraystretch}{0.5} 

\begin{tabular}{cccccccc}
\includegraphics[width=0.64in]{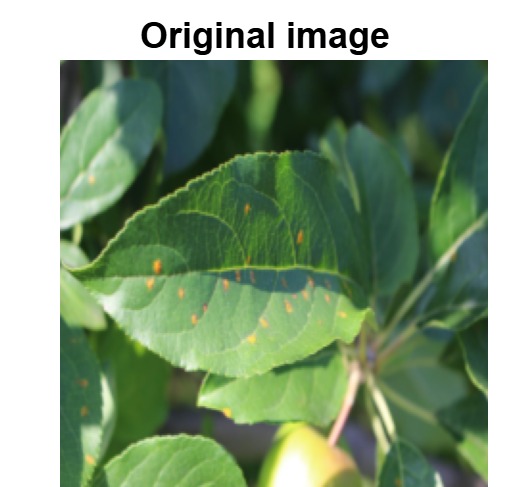} &
\includegraphics[width=0.5in]{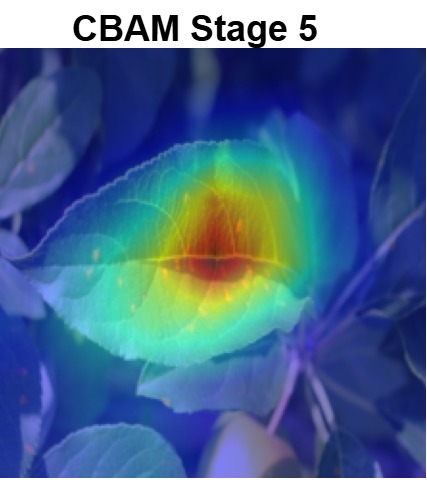} &
\includegraphics[width=0.5in]{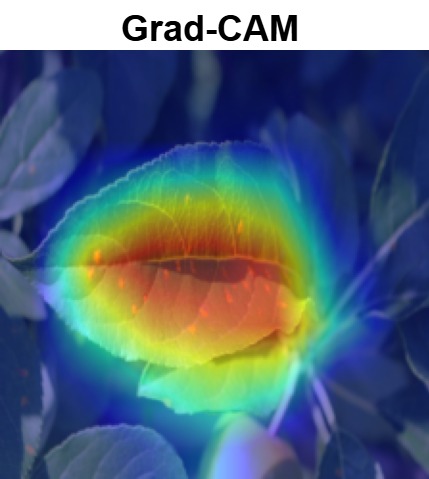} &
\includegraphics[width=0.5in]{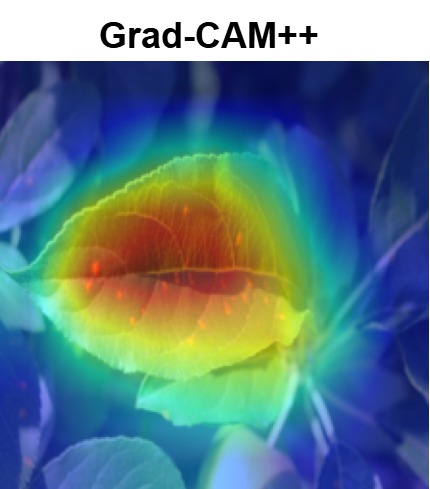} &
\includegraphics[width=0.51in]{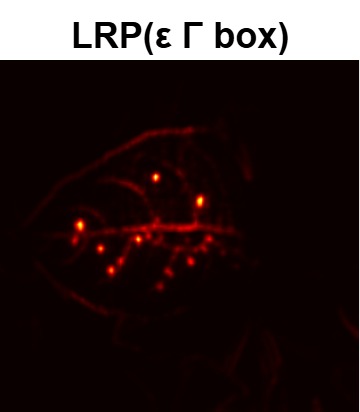} &
\includegraphics[width=0.51in]{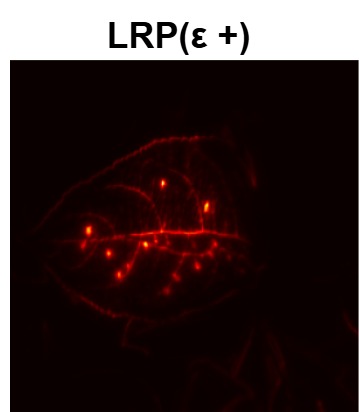} &
\includegraphics[width=0.49in]{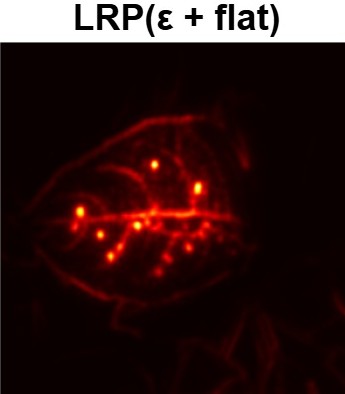} &
\includegraphics[width=0.5in]{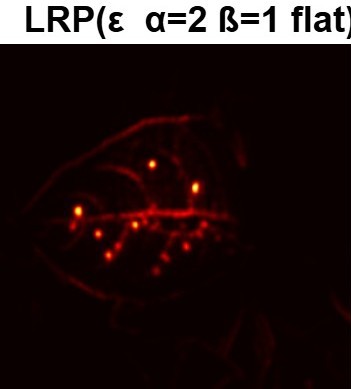} \\

\includegraphics[width=0.5in]{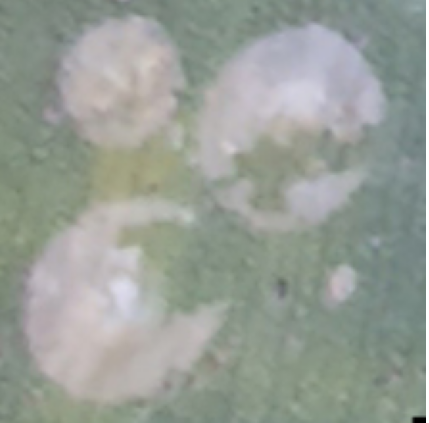} &
\includegraphics[width=0.5in]{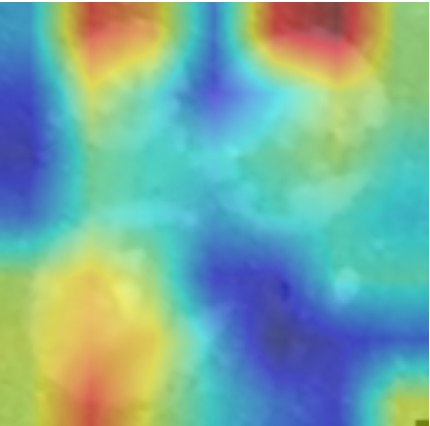} &
\includegraphics[width=0.51in]{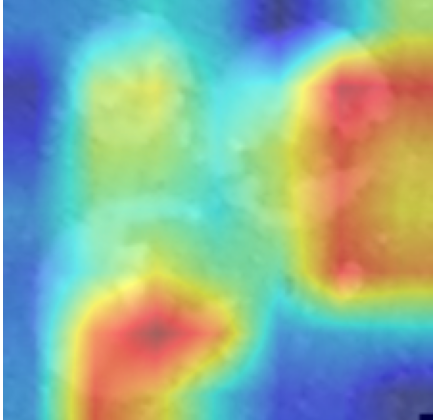} &
\includegraphics[width=0.5in]{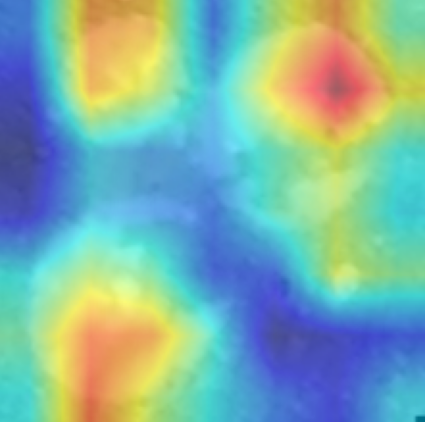} &
\includegraphics[width=0.5in]{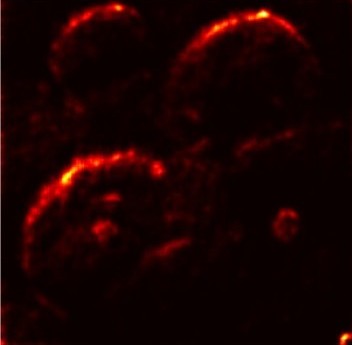} &
\includegraphics[width=0.5in]{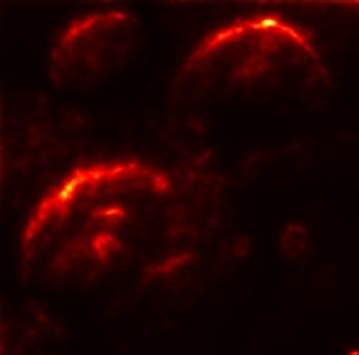} &
\includegraphics[width=0.5in]{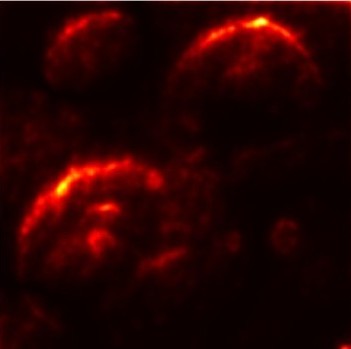} &
\includegraphics[width=0.5in]{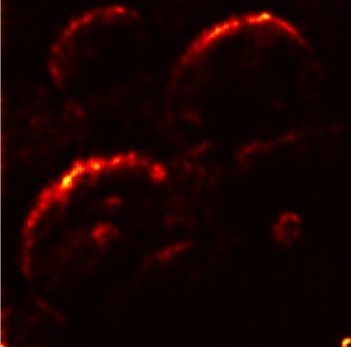} \\

\includegraphics[width=0.51in]{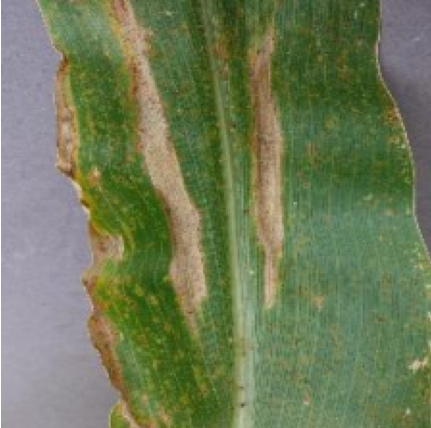} &
\includegraphics[width=0.5in]{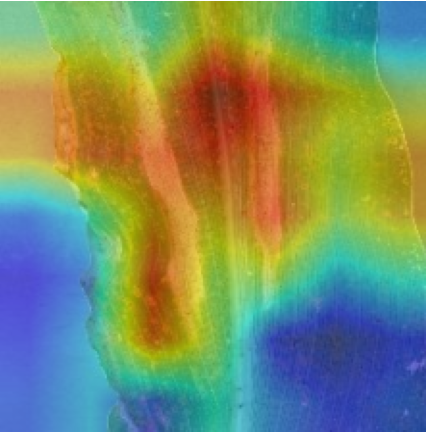} &
\includegraphics[width=0.51in]{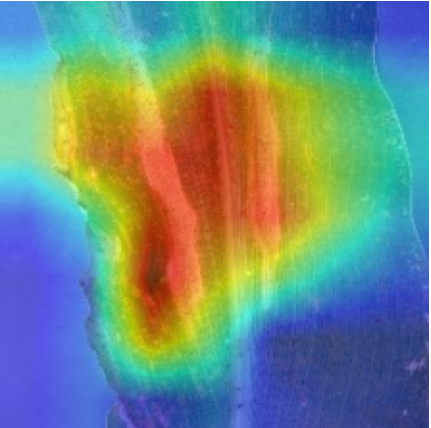} &
\includegraphics[width=0.51in]{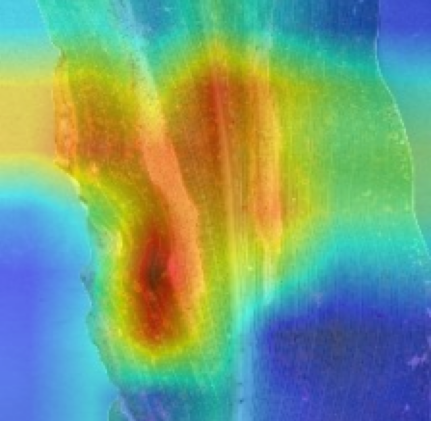} &
\includegraphics[width=0.52in]{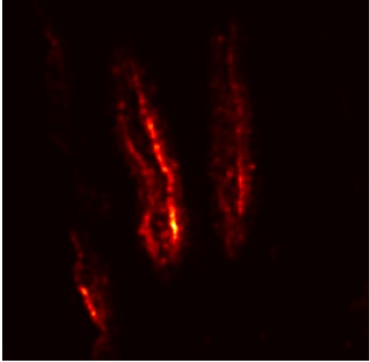} &
\includegraphics[width=0.52in]{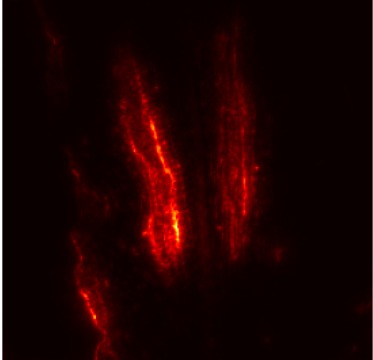} &
\includegraphics[width=0.5in]{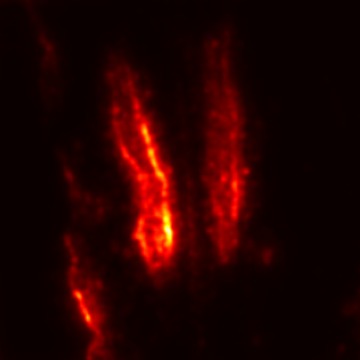} &
\includegraphics[width=0.5in]{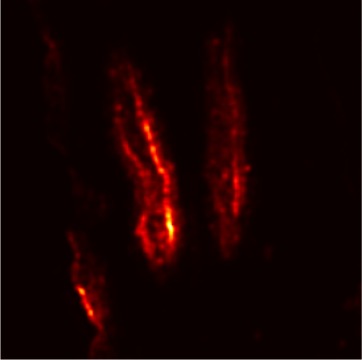} \\

\includegraphics[width=0.5in]{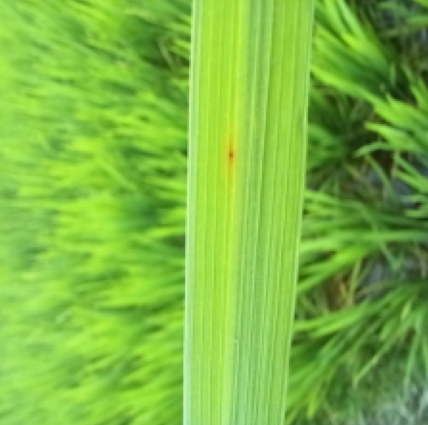} &
\includegraphics[width=0.5in]{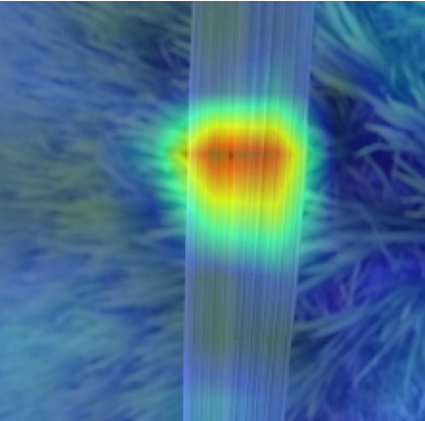} &
\includegraphics[width=0.5in]{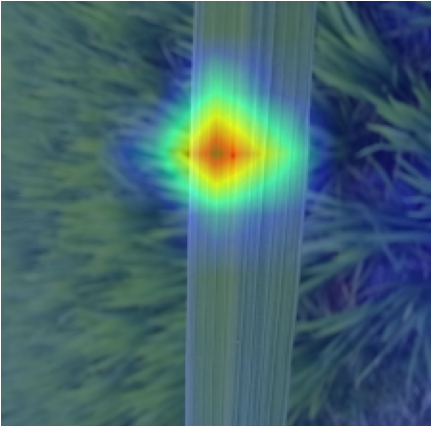} &
\includegraphics[width=0.5in]{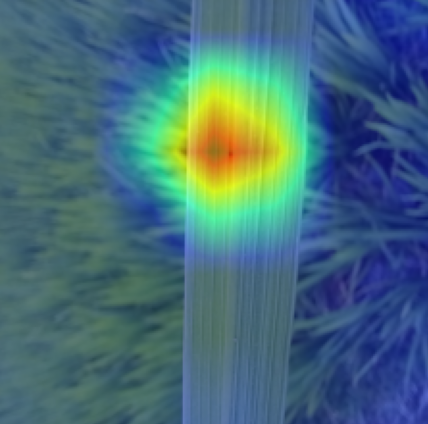} &
\includegraphics[width=0.5in]{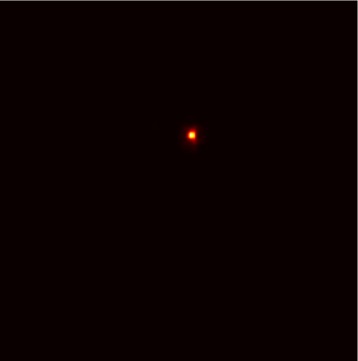} &
\includegraphics[width=0.5in]{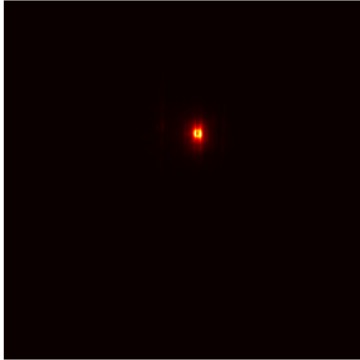} &
\includegraphics[width=0.5in]{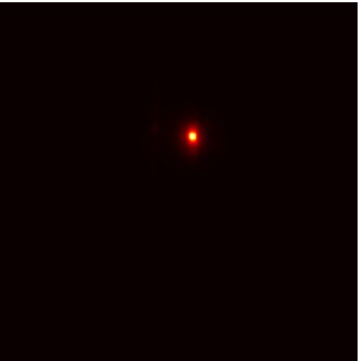} &
\includegraphics[width=0.5in]{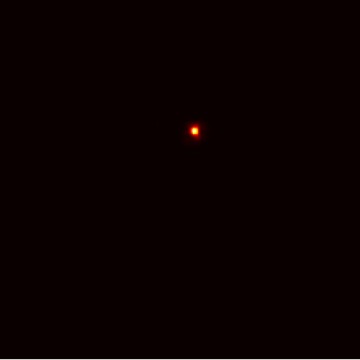} \\
\end{tabular}

\caption{Explainability analysis for other crop leaves used in the experiments. Top to bottom row represents leaves of Apple, Embrapa (Coconut), Maize, and Rice.}
\label{fig:lrp_grid}
\end{figure*}

We have selected grape leaf dataset from plant village dataset for detailed explainability analysis of our achieved results to validate the model's interpretability. In the Fig. \ref{fig:grape_leaf_explanations_2}, CBAM attention maps generated across all the five convolutional stages inserted in the  proposed CBAM-VGG16 model for grape leaf classification are shown. These maps provide insight into how the model attends to informative regions at various depths of the network. 
Overall, these visualisations demonstrate how CBAM refines feature learning progressively across the network, resulting in improved focus, interpretability, and diagnostic accuracy.

Fig.~\ref{fig:grape_leaf_explanations_1} displays a visual comparison of Grad-CAM and Grad-CAM++ heat maps for various leaf samples across different disease classes. Grad-CAM computes gradients reaching the last convolutional layers to determine which input locations most affected the model's decision-making. However, it typically produces relatively coarse, blob-like heatmaps. In contrast, Grad-CAM++ refines this by incorporating second-order gradients, resulting in more spatially accurate and class-discriminative attention maps.

 Fig.~\ref{fig:lrp_grid_1} illustrates the class-wise attribution heatmaps generated for the CBAM-VGG16 model using a comprehensive suite of LRP techniques. 
 The LRP family consistently produces sparse yet discriminative saliency maps, aligning well with regions exhibiting disease-related symptoms. Among them, \textit{Epsilon Plus Flat} and \textit{Epsilon Alpha2 Beta1 Flat} stand out for their capacity to sharply localize high-relevance areas, typically corresponding to lesions, necrotic margins, or discolored patches symptomatic of disease. These variants effectively suppress irrelevant background and vein structures, enhancing focus on pathologically significant textures.
 
We have shown in  Fig. \ref{fig:lrp_grid} the explainable visualization of proposed method on some other disease of other dataset used in this work. These visualisations also exhibit the same properties as observed in the detailed explainability analysis for grape leaf dataset.

\subsection{Feature Analysis}

\begin{figure}[t]
    \centering
    \includegraphics[width=0.6\linewidth]{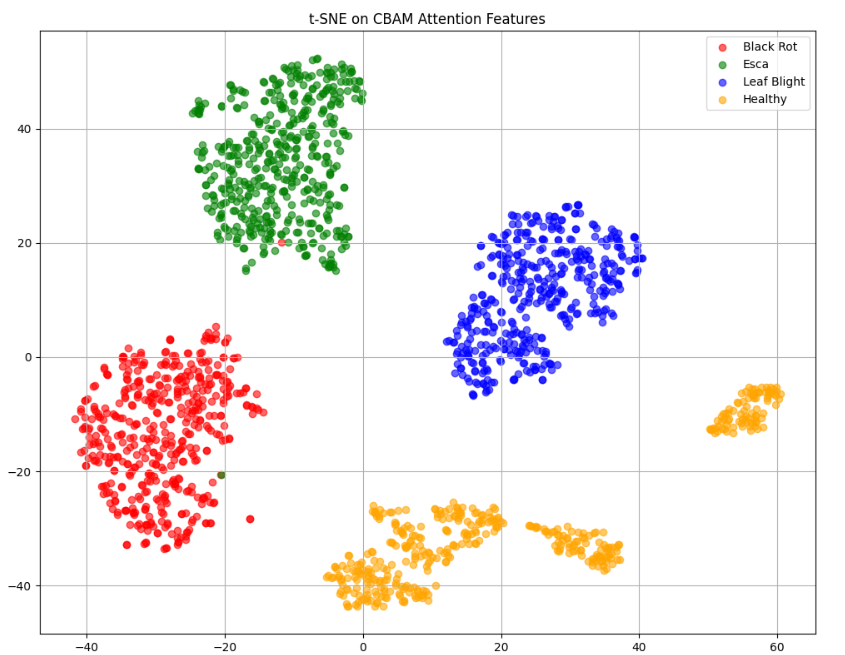}
    \caption{Visualization of extracted features using t-SNE plots.}
    \label{fig:tsne_vgg_cbam}
\end{figure}

\begin{figure}[t]
    \centering
    \includegraphics[width=0.56\linewidth]{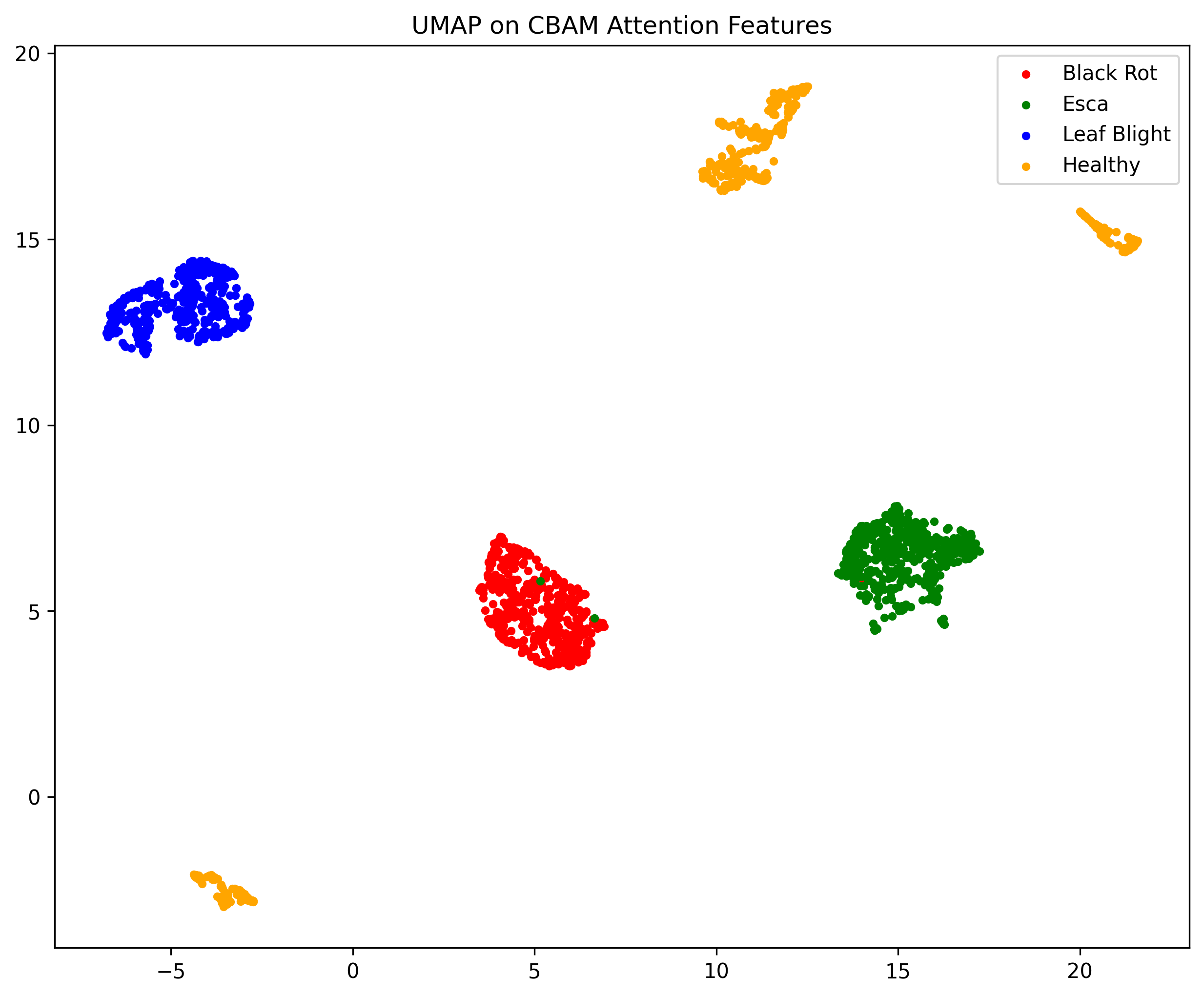}
    \caption{Visualization of extracted features using UMAP plots.}
    \label{fig:UMAP}
\end{figure}

To evaluate the discriminative capability of the CBAM-enhanced VGG16 model, we employed t-SNE and UMAP to project high-dimensional feature representations into a 2-D space. Fig.~\ref{fig:tsne_vgg_cbam} and Fig. \ref{fig:UMAP} presents the t-SNE and UMAP plots on the four grape leaf classes: \textit{Esca}, \textit{Healthy}, \textit{Leaf Blight}, and \textit{Black Rot}. It can be interpreted from the t-SNE figure that our proposed model is providing a clear distinguishable cluster separation with minimal class overlap, indicating better feature discrimination. In the UMAP plot, class boundaries are
clearly defined, with each category forming dense
and distinct clusters. These findings demonstrate how well CBAM works to improve the VGG16 backbone's representational quality, resulting in improved class-wise separability and model interpretability.




\section{Conclusion}
\label{conclusion}
The study in this work provides an interpretable approach for detection of plant leaf diseases. By incorporating attention modules at each convolutional stage, the model not only enhances disease-specific feature extraction but also offers improved performance across five public datasets. The performance on five different datasets exhibits the genralizability of our proposed solution for any other crop. In order to evaluate the model's interpretability, we carried this in-depth evaluation using different attribution techniques. The qualitative study highlighted that LRP variants produced the most visually clear, localized, and class-discriminative heatmaps with minimal noise. Future directions include refining the attention mechanism to enhance class-awareness and reduce interpretability gaps, exploring global or transformer-inspired attention integration for better contextual representation, and conducting human-in-the-loop evaluations to assess the trust and reliability of visualisations.

\backmatter









\subsubsection*{Author Contribution} Balram singh has implemented the work and written the first draft of the work. Ram Prakash Sharma has helped to formulate the problem, develop the solution, and proofread the manuscript. Somnath Dey has improved the problem solution with proofreading of the manuscript.

\subsubsection*{Funding} The authors declare that no funding was received for this research.

\subsubsection*{Data Availability} This study has not generated any datasets during the
current study.

\section*{Declarations}
\subsubsection*{Conflict of interest} The authors declare that they have no conflict of interest.

{\footnotesize
\bibliography{references}}

\end{document}